%% file: main.tex
\documentclass{article}

\PassOptionsToPackage{numbers, compress}{natbib}

\usepackage[preprint]{neurips_2025}

\input{preamble}

\title{DropCluster: \\Structured Dropout for Convolutional Networks
}

\author{
Liyan Chen
\textsuperscript{1} 
\hspace{15pt}
Philippos Mordohai 
\textsuperscript{1}
\hspace{15pt}
Serg\"ul  Ayd\"ore
\textsuperscript{2}
\And
\textsuperscript{1}
    Stevens Institute of Technology
\hspace{15pt}
\textsuperscript{2}
    Amazon Web Services
\hspace{15pt}
}

\begin{document}

\maketitle

\input{sec/Abstract}

\section{Introduction}\label{sec:introduction}
\input{sec/Introduction}

\section{Related Work}\label{sec:related_work}
\input{sec/Related_work}

\section{Motivation}\label{sec:motivation}
\input{sec/Motivation}

\section{Method}\label{sec:dropcluster} 
\input{sec/Method}

\section{Experimental Results}\label{sec:experiment}
\input{sec/Experiments}

\section{Conclusion}\label{sec:conclusion}
\input{sec/Conclusion}

{
\small
\bibliographystyle{ieee_tran}
\bibliography{mybib}
}

\clearpage
\newpage
\appendix
\section*{Appendix}
\input{sec/Supplement}




\end{document}

%% file: preamble.tex
\usepackage{amssymb}
\usepackage{amsthm}
\usepackage{amsmath}
\usepackage{array}
\usepackage{algorithm}
\usepackage{algorithmic}
\usepackage{adjustbox}
\usepackage{bm}
\usepackage{booktabs}
\usepackage{capt-of}
\usepackage{graphicx}
\usepackage{hyperref}
\usepackage[figuresright]{rotating}
\usepackage{threeparttable}
\usepackage[table]{xcolor}

\usepackage[utf8]{inputenc} 
\usepackage[T1]{fontenc}    
\usepackage{url}            
\usepackage{amsfonts}       
\usepackage{nicefrac}       
\usepackage{microtype}      
\usepackage{enumitem}

%% file: sec/Abstract.tex
\begin{abstract}
Dropout as a common regularizer to prevent overfitting in deep neural networks has been less effective in convolutional layers than in fully connected layers. This is because Dropout drops features randomly, without considering local structure. When features are spatially correlated, as in the case of convolutional layers, information from the dropped features can still propagate to subsequent layers via neighboring features. To address this problem, structured forms of Dropout have been proposed. A drawback of these methods is that they do not adapt to the data. In this work, we leverage the structure in the outputs of convolutional layers and introduce a novel structured regularization method named DropCluster. Our approach clusters features in convolutional layers, and drops the resulting clusters randomly during training iterations. Experiments on CIFAR-10/100, SVHN, and APPA-REAL datasets demonstrate that our approach is effective and controls overfitting better than other approaches.
\end{abstract}

%% file: sec/Introduction.tex
Convolutional neural networks (CNNs) have become a foundational tool in several scientific fields, such as computer vision. Regularization is needed during CNN training to ensure that the model generalizes well to unseen data, especially when the training set is small. Achieving good generalization, however, is crucial for applications such as medical diagnostics, self-driving cars, or face recognition. 

Dropout is a widely used regularization approach to reduce generalization error \cite{JMLR:v15:srivastava14a} and mitigate overfitting in the context of overparameterized models. It randomly drops features\footnote{We adopt the term \textit{features} from DropBlock \cite{ghiasi2018dropblock}, which is the most similar approach to ours, to denote neurons/units of the network, and the term \textit{feature maps} to emphasize the spatial structure of network layers.} in hidden layers during training, and can be interpreted as a way of ensembling thinned networks at test time. 

Despite the smaller number of parameters in convolutional layers compared to fully-connected ones, CNNs still encounter overfitting issues, as evidenced by various studies. Mostafa et al.~\cite{mostafa2021visualizing} provided an experiment that further emphasizes the tendency of convolutional networks to overfit when neurons learn highly similar features. Srivastava et al.~\cite{JMLR:v15:srivastava14a} highlighted that dropout remains effective in combating overfitting in convolutional layers by introducing noise to inputs for higher fully connected layers. Structured dropout, as explained by Tompson et al.~\cite{tompson2015efficient}, aims to enhance generalization by preventing strong correlations among neighbor features. These references collectively affirm that CNNs are prone to overfitting, and dropout-style regularization proves beneficial in addressing this issue.

In this paper, we introduce a novel structured regularization method for convolutional layers, called DropCluster. DropCluster learns inherent structure within feature maps and leverages this knowledge to drop features more effectively, thereby alleviating overfitting in CNNs.

Like SpatialDropout~\cite{tompson2015efficient} and DropBlock \cite{ghiasi2018dropblock}, DropCluster aims to suppress the spatial correlation between adjacent features. Since the clusters in the feature maps have arbitrary shapes, dropping features in square blocks, as in DropBlock, may include features from different clusters, or leak information through the surviving members of the clusters. On the other hand, dropping the entire channel, as in SpatialDropout, suppresses vital information for recognition. 

In contrast, DropCluster adapts to the observed structure of the data and selectively drops specific clusters within each channel, hence striking a balance between information preservation and spatial decorrelation.
Empirical results demonstrate that abstracting prototypes or visual concepts from a few examples, rather than starting from scratch, is effective for improving the classification of new data~\cite{fei2006one, lake2011one,snell2017prototypical}. A recent study~\cite{liu2023dropout} also suggests that applying dropout before a certain iteration and then disabling it prevents underfitting, while the opposite sequence can help alleviate the risk of overfitting.

Avoiding overfitting in deep networks can be particularly challenging in few-shot learning \cite{ravi2016optimization, kaiser2017learning}, and becomes even more crucial in domains where data acquisition is prohibitively expensive, such as astronomy, genomics, chemistry, and medical imaging\cite{fan2006statistical,10002015global}. Most benchmark datasets,  however, are large, and may not reflect the overfitting problem that occurs in real-world applications. In order to mimic settings favoring overfitting, we run experiments on smaller datasets and neural networks, and provide performance comparisons between DropCluster and other dropout-inspired regularizers. 
Our results in Section \ref{sec:experiment} show that DropCluster outperforms alternative approaches in classification and regression tasks when overfitting is a problem.

In summary, our contributions are: (a) a novel structured dropout method for convolutional layers, named DropCluster, which leverages data-driven structure; (b) a statistical measure to assess clustering tendency in 2D data, and evidence that the spatial structure of convolutional layers can be learnt during training; and (c) empirical results showing that DropCluster prevents overfitting in CNNs for both classification and regression. Our code will be open-sourced after the paper is accepted.

%% file: sec/Related_work.tex
Dropout \cite{JMLR:v15:srivastava14a} is a popular and simple regularization approach to prevent overfitting in neural networks. 
Empirical studies show that structured variants of Dropout such as \cite{tompson2015efficient,huang2016deep,ghiasi2018dropblock,DBLP:conf/aaai/HouW19,dai2019batch, DBLP:conf/ijcai/ChenNLT20,DBLP:conf/icml/ShiZDZMW20,DBLP:journals/ijcv/ZuninoBMZSM21,DBLP:conf/aaai/PhamL21} attain better performance than the original Dropout on convolutional networks. 

SpatialDropout \cite{tompson2015efficient} randomly drops channels of the network, while StochasticDepth \cite{huang2016deep} randomly removes a subset of layers during training while maintaining a full-depth network at test time. Different from regular dropout methods that zero out a subset of features, DropConnect \cite{pmlr-v28-wan13} randomly selects a subset of weights within the network to be zero. Both methods are informed by the network architecture but not by patterns observed during model training. 
Instead of totally random selection, Huo and Wang~\cite{DBLP:conf/aaai/HouW19} introduce Weighted Channel Dropout, which keeps the highly activated channels with high probability. %
Conversely, SelectScale \cite{DBLP:conf/ijcai/ChenNLT20} drops the most important features based on their values after activation layers. Informative Dropout \cite{DBLP:conf/icml/ShiZDZMW20} aims to suppress the texture-bias in the data and zeros out with higher probability outputs for input patches containing less information. By filtering the texture, this approach improves the robustness of the model against distribution shift and adversarial perturbations. 

The approach that is most closely related to ours is DropBlock \cite{ghiasi2018dropblock}, which randomly drops small square regions in feature maps. Unlike conventional Dropout, DropBlock improves training of convolutional layers because selecting contiguous blocks leverages the spatial correlation found in feature maps. This prevents information from dropped features leaking into training via the features' correlated neighbors. Pal et al.~\cite{DBLP:conf/cvpr/PalLVH20} provide a theoretical explanation for the empirically superior performance of DropBlock. (See Section \ref{sec:objective}.)
Similarly, Batch DropBlock \cite{dai2019batch} randomly drops the same region of all input feature maps in a batch to reinforce attentive feature learning of local regions. Corrdrop~\cite{zeng2020corrdrop} integrates DropBlock and SelectScale, enhancing the approach by refining the dropout probabilities of each block based on the mean feature orthogonality with other units.
However, the shape of the dropped regions in DropBlock-like methods is fixed despite the potential irregularity in the local structure of feature maps. Indeed, our observations in Section~\ref{sec:motivation} reveal variations in the shapes of correlated regions within feature maps across different channels. Our algorithm utilizes these varying shapes directly, dropping precisely the features that are highly correlated with each other. Divergent from DropBlock and the other methods above, AutoDropout \cite{DBLP:conf/aaai/PhamL21} can be viewed as data augmentation applied on CNNs and transformers. While dropping out a continuous region such as rectangles in the hidden layers, AutoDropout also applies geometric transformations like rotation and shearing to the dropout patterns.

Relevant research in adjacent areas of machine learning includes the work of \cite{liu2023patchdropout}, which applies structure dropout on vision transformers to improve their generalization ability. Lee et al. \cite{lee2022self} proposed to combine dropout with self-knowledge distillation to improve calibration performance, adversarial robustness, and out-of-distribution detection. %
Lin et al. \cite{lin2023explore} study the use of dropout to aid few-shot learning and conclude that dropout should be applied in the pre-training phase, which enhances the model's generelizability, but not in the fine-tuning phase in target domain.

Common deep architectures like ResNet and U-Net contain residual connections, which limits the performance of applying dropout methods. However, in many related works mentioned above, the improved performance on residual networks with the dropout methods also comes with a modification of network architecture. Zhang et al. \cite{zhang2019confidence} also demonstrate that due to the skip connections, there is still an information flow through the network even after dropping out an entire layer.
Veit et al. \cite{veit2016residual} provide a study that the paths with skip connections contribute to gradient during training. 
Kim et al. \cite{kim2023use} propose that applying dropout after the last batch normalization but
before the last weight layer in the residual branch could increase the performance of residual networks.
In this study, our primary focus is to examine the efficacy of DropCluster in addressing overfitting specifically in convolutional layers. To streamline the variables and eliminate the influence of residual connections, our experiments are exclusively conducted on lightweight CNNs.

In classical Dropout, the drop probability of each neuron is equal, and can be viewed as introducing multiplicative Bernoulli noise to prevent overfitting. 
There are, however potential benefits to be gained by adapting the drop probabilities.
Several researchers propose to alter the drop probability over different units or iterations instead of using a constant value. Approaches applicable to convolutional layers include those of \cite{ba2013adaptive,dodballapur2020automatic,fan2021contextual}, while other researchers have investigated similar techniques on different layers or architectures, such as transformers \cite{li2022dropkey}. We consider the automatic adaptation of drop probabilities a direction for future work.

%% file: sec/Motivation.tex
\begin{figure*}[t]
\begin{center}
\adjustbox{width=\textwidth}{
\begin{tabular}{ccc}
\includegraphics[width=0.3\columnwidth]{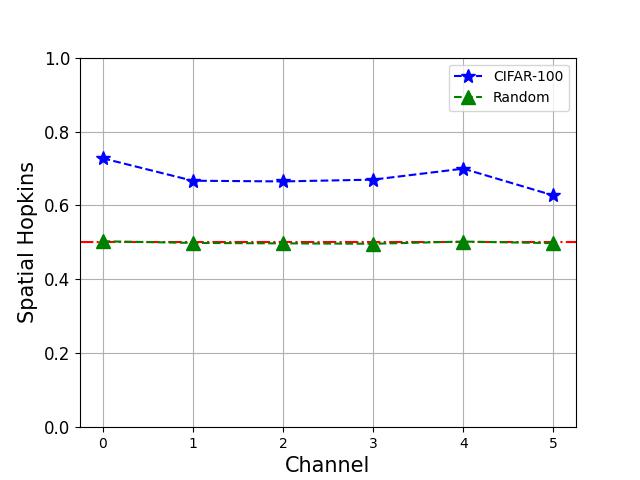} &
\includegraphics[width=0.3 \columnwidth]{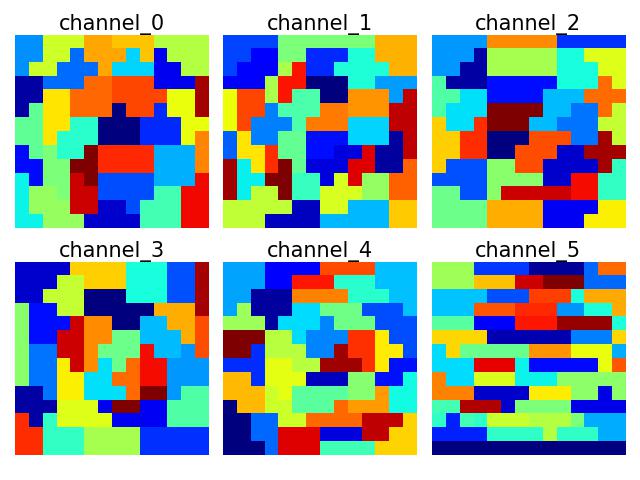} &
\includegraphics[width=0.3 \columnwidth]{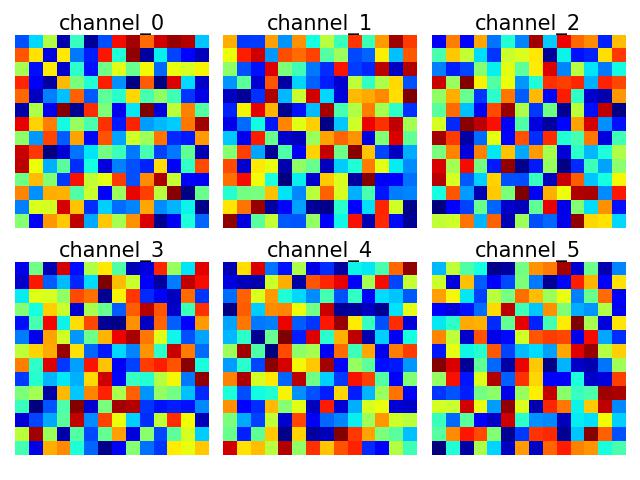} \\
(a) Spatial Hopkins statistics &
(b) Clusters of structured features &
(c) Clusters of random features \\
\end{tabular}
}
\end{center}
\captionof{figure}{
(a) Spatial Hopkins statistic for six channels with real (blue curve) and random inputs (green curve) in \textit{b} and \textit{c}, respectively. (b) Clusters from the first convolutional layer outputs from a LeNet-5 network trained on CIFAR-100 with real inputs. (c) Similar to b for random inputs.
}
\label{fig:clustering_tendency}
\end{figure*}

In this section, we present evidence that output feature maps of convolutional layers exhibit \textit{clustering tendency} (see Fig.~\ref{fig:clustering_tendency}) to highlight the need for DropCluster. We demonstrate clustering tendency qualitatively by visualizing clusters, and quantitatively by computing clustering statistics. We extend the Hopkins statistic \cite{banerjee2004validating}, 
which measures the degree to which clusters exist in data, to introduce a new statistic appropriate for spatial data, such as images. 

\subsection{Qualitative Evidence}
We train a LeNet-5 network \cite{lecun1989backpropagation} on the CIFAR-10/100 dataset \cite{krizhevsky2009learning}, then compute clusters in each channel given a mini-batch of outputs from the first convolutional layer. We observe that clusters in each feature map change rapidly during the early stages of training. Therefore, we do not attempt clustering until the model stabilizes.

To visualize the computed clusters, we train LeNet-5 on CIFAR-10 for 120 epochs. When overfitting is observed (the validation loss increases), we cluster the outputs of the first convolutional layer. The two images at the top of the first column in Fig.~\ref{fig:Pipline_for-DropCluster} are the original input samples (horse and bird). The black-and-white images below them are the feature maps of these two samples after the first convolutional layer. The second column in Fig.~\ref{fig:Pipline_for-DropCluster} shows the clusters computed for each channel from a batch of images. It is clear that the relatively large feature maps produced by the first convolutional layer contain recognizable features with obvious connections to the raw input images. Furthermore, the correlated groups of features identified by clustering are irregularly shaped. By dropping entire clusters, our approach suppresses spatially correlated information effectively. DropBlock, on the other hand, is limited to dropping square blocks that straddle clusters. 

\subsection{Spatial Clustering Statistics}
Here, we present quantitative evidence for the presence of clusters in feature maps. We first discuss the Hopkins statistic \cite{banerjee2004validating}, a measure for quantifying the presence of clusters in data. Then, we show a novel adaptation of the Hopkins statistic 
that is applicable to 2D data. 

In the original paper \cite{banerjee2004validating}, computation of the Hopkins statistic begins by generating $m$ random points uniformly distributed in a sampling window, and also sampling $m$ actual data points from the collection of $n$ data points, $(m << n)$. We measure the distance from the actual points to the nearest neighbor in both sets. 
Let $u_i$ denote the nearest neighbor distance from artificial point $i$ to the entire data set, and $w_i$ denote the nearest neighbor distance from sampled actual point $i$ to the entire data set. Then the Hopkins statistic is:

\begin{equation}
H := \frac{\sum_{i=1}^m u_i}{\sum_{i=1}^m u_i + \sum_{i=1}^m w_i}.
\end{equation}

When the actual data points are approximately uniformly distributed, they will have similar distances to the randomly generated artificial points, and $H$ will be close to $0.5$, implying the lack of clustering tendency. %
If $H$ is close to $1$, we conclude that the data exhibit strong clustering tendency. 
\begin{figure*}[t]
	\centering
	\includegraphics[width=\textwidth]{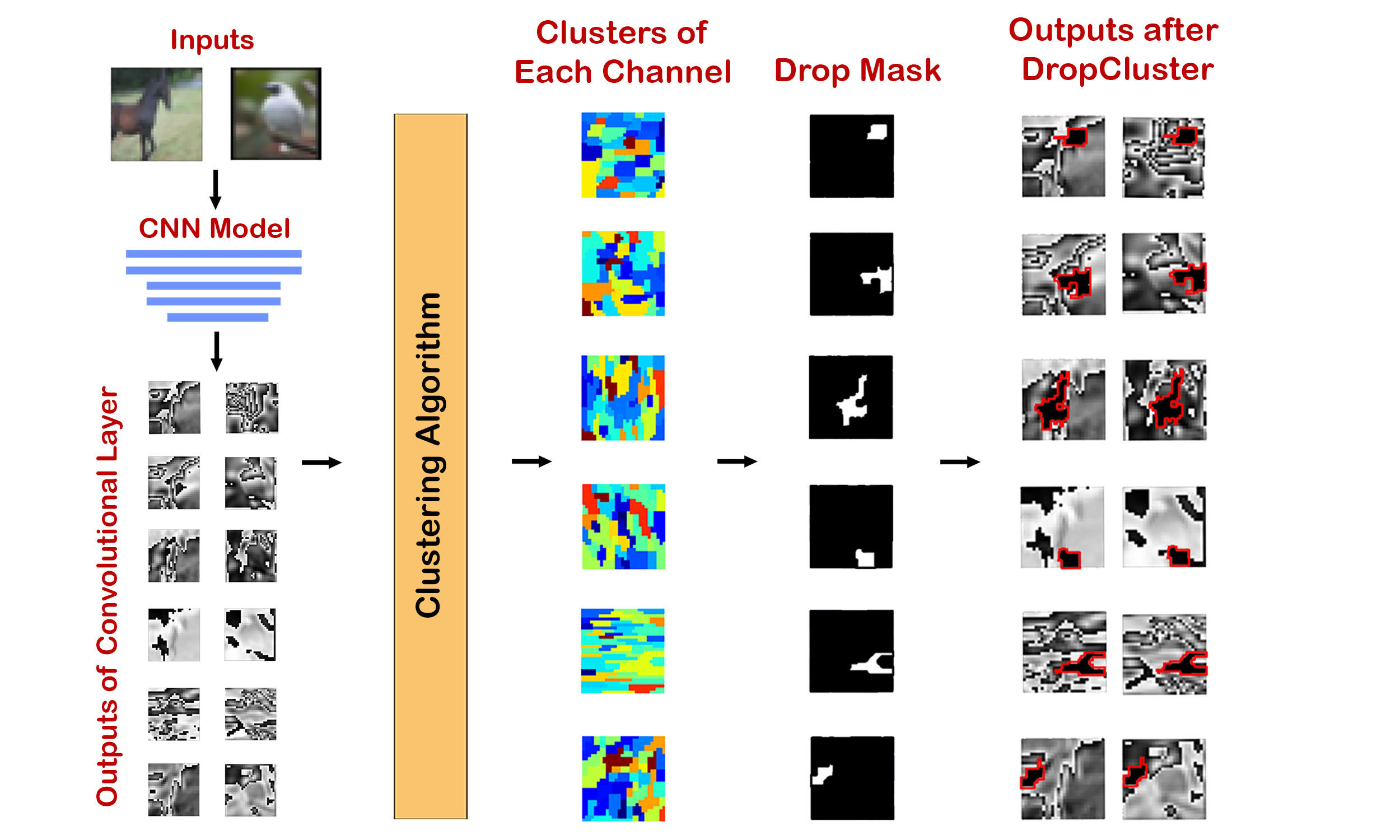}	
	
	\caption{An illustration of a CNN trained with DropCluster using the output of the first convolutional layer of LeNet-5 trained on the CIFAR-10 dataset to visualize our method. DropCluster starts by training a CNN without dropout layers, computes clusters on the resulting activations,
	and then applies dropout on the clusters.}
	\label{fig:Pipline_for_DropCluster}
\end{figure*}

The Hopkins statistic, however, is not applicable to image data because it does not take the spatial locations of pixels into account. 
Therefore, we propose a new clustering tendency statistic \textit{Spatial Hopkins} based on the difference of each pixel's intensity from its neighbors. When there is spatial structure in the data, each pixel's intensity value will be close to its neighbors' values. Based on this observation, we sample $m$ pixel locations $(x_i, y_i)$ uniformly sampled from $x_i \in \{2, \cdots, X-1 \}$ and $y_i \in \{2, \cdots, Y-1\}$, $i \in 1, \cdots, m$ for an image $I$ of size $X \times Y$. 
Similar to the Hopkins statistic, we choose $m << X \times Y$. Then we compute the average $L_2$ distance between the value of the pixel at each location and its $8$ neighbors as:

\begin{equation}
w_i = \frac{1}{8} \sum_{u, v} \|I(x_i, y_i) - I(u, v)\|_2
\end{equation}
where $(u, v)$ span the 8-neighborhood of  $(x_i, y_i)$.
Next, we sample $m$ additional pixel locations, $(q_1, r_1), \cdots, (q_m, r_m)$ again following a uniform distribution in the image, and compute the average distance between the intensities of $(q_i, r_i)$ and the neighbors of $(x_i, y_i)$:
\begin{equation}
z_i = \frac{1}{8} \sum_{u, v} \| I(q_i, r_i) - I(u, v) \|_2.
\end{equation}
The set of points ${(u,v)}$ are the neighbors of $(x_i, y_i)$ in both equations. Finally, similar to the Hopkins statistic, we compute the Spatial Hopkins statistic for a given image as:
\begin{equation}
S := \frac{\sum_{i=1}^m z_i}{\sum_{i=1}^m z_i + \sum_{i=1}^m w_i}.
\end{equation}

The intuition is that, if there is no
structure, the average distance between a pixel to its neighbors will be equal to the average distance to a random pixel, i.e. $w_i$ and $z_i$ would be approximately equal, and the Spatial Hopkins statistic will have values near $0.5$.

Here, we treat a layer of the network as an image. As the descriptor of each feature, we use its activation values over a mini-batch of input images. That is, a feature is associated with the $b$ activation values generated by the forward pass of a mini-batch with $b$ images. Clustering is performed separately in each channel. In Fig.~\ref{fig:clustering_tendency}, we report the average Spatial Hopkins values across a mini-batch for each channel from the outputs of the first convolutional layer of CIFAR-100 with a LeNet-5 network trained until overfitting begins. It can be seen that all channels have Spatial Hopkins values larger then $0.5$ indicating the presence of structure. We also provide Spatial Hopkins values for random inputs as baseline, which are unsurprisingly around $0.5$.

The variability in cluster shapes found in feature maps indicates the need for data-driven regularization that adapts to the shape of clusters.
We present such an approach in the next section.

%% file: sec/Method.tex
In this section, we present DropCluster in more detail. Following the analysis of Liu et al. \cite{liu2023dropout}, the clusters are learned after some epochs of pre-training without dropout. We then randomly drop clusters of correlated features in each channel in subsequent training iterations. Similar to standard Dropout, no neurons are dropped in testing. We give the flowchart of our approach in Fig.~\ref{fig:Pipline_for-DropCluster}, which includes four steps: (a) Pre-training a No-drop model; (b) Discovering the number of clusters; (c) Computing clusters; (d) Applying dropout on the clusters.

\subsection{Pre-training a No-drop Model}
We first train a CNN model without dropout, referred as the \textit{No-drop} model, until it begins overfitting. Both training and validation losses change rapidly during the first several iterations, indicating changes in the learnt structure. 
The goal is to obtain a pre-trained model that has learned structure from the data and is stable, 
but is not too specific to the training set.
Thus, we save the pre-trained model where training loss continues to decrease and validation loss starts to increase.

\subsection{Discovering the Number of Clusters}
To learn the structure of the convolutional outputs, we need to discover 
the number of clusters. We first apply a clustering algorithm such as k-means \cite{lloyd1982least} to obtain $k$ clusters from the outputs of a convolutional layer for a range of values of $k$, and
use the Silhouette Coefficient to determine the best $k$ for each channel. (See Algorithm \ref{alg:compute_clusters} in Appendix \ref{sec:compute_cluster}.)

\paragraph{The Silhouette Coefficient} \cite{rousseeuw1987silhouettes} was proposed to evaluate clustering validity based on cluster tightness and separation. For each sample $i$ in cluster $A$, let $a(i)$ be the mean intra-cluster distance, and $b(i)$ be the mean distance from $i$ to samples in the cluster that is nearest to $A$. 

The Silhouette Coefficient of element $i$ is defined as
\begin{equation}
    s(i) = \frac{a(i)-b(i)}{\max \left\{a(i),b(i)\right\}}.
\label{silhouette}
\end{equation}

The average Silhouette Coefficient across samples is computed for every cluster using Eq.~\ref{silhouette}. Since a higher Silhouette Coefficient indicates higher quality of partitioning, we select the value of $k$ that maximizes the average Silhouette Coefficient over all clusters.

\subsection{Computing Clusters}
\label{sec:feature_assigment}
For each channel, we compute the cluster masks using the best $k$ picked in the previous step. We cluster the features based on the activations over a batch of $b$ samples using 
k-means \cite{lloyd1982least}. (Note that other clustering algorithms could be used at this step as well.) Each cluster contains several features, and each feature is represented by a sequence of $b$ activations, one for each image in the batch. 
The computed clusters are encoded by  
$\boldsymbol{\Phi} \in \mathbb{R}^{k \times d}$, the feature assignment matrix 
that represents the $k$ clusters of the $d$ features such that
\begin{equation}
\boldsymbol{\Phi} = \left[\begin{array}{cccc} 1 \dots  1 & 0 \dots 0 & \dots & 0 \dots 0 \\ 0 \dots 0 & 1 \dots 1 & \dots & 0 \dots 0 \\
\vdots & \vdots & \vdots & \vdots \\ 0 \dots 0 & 0 \dots 0 & 0 \dots 0 & 1 \dots 1 \end{array} \right]
\end{equation}
after an appropriate permutation of the features.
Rows of $\boldsymbol{\Phi}$ correspond to the clusters, and columns to the features.
Each feature is assigned exactly to one cluster.
This and the previous step aim to obtain the best feature assignment matrix  $\boldsymbol{\Phi}_{best}$. They are performed only \textit{once} during the entire training process. 

\subsection{Applying Dropout on the Clusters}
In the last step, we continue training the CNN model and apply dropout on the clusters. For each training iteration, we randomly sample dropout variables for each channel based on the drop probability. We create drop masks for the corresponding clusters of each channel and then apply the masks to the outputs of the channel. The drop masks at each iteration and for each channel should be different in general. See the ``Drop Mask'' column in Fig.~\ref{fig:Pipline_for-DropCluster} and Algorithm~\ref{alg:dropcluster}. 

\begin{algorithm}[h]
   \caption{DropCluster
   }

   \textbf{Input:} 

   \hspace{\parindent} $\mathbf{X} \in \mathbb{R}^{b \times c \times d}$, output activations of the convolutional layer, where $b$ is the mini-batch size, $c$ is the number of channels and $d$ is the size of the feature map at each channel.

   \hspace{\parindent} Set of $\mathbf{\Phi}_{best}^{(i)} \in \{0,1\}^{k \times d}$, best feature assignments for channel $i$

   \hspace{\parindent} $\theta$, the keep probability

   \hspace{\parindent} $k$, the number of clusters

   \textbf{Output:} 

   \hspace{\parindent} $\bar{\mathbf{X}}$: the masked outputs

   \label{alg:dropcluster}
\begin{algorithmic}[1]
\STATE Initialize mask matrix to ones: $\mathbf{M} = \mathbf{1}_{c \times d}$
\IF {$mode == Training$}
\FOR{channel $i=1$ {\bfseries to} $c$}
	\STATE Sample cluster dropout variables $\mathbf{z}^{\left(i\right)} \in \mathbb{R}^k$, \\  
 where $\mathbf{z}^{\left(i\right)}_j \sim \mbox{Ber}\left(\theta\right)$
		\STATE $\mathbf{M}[i,:] \leftarrow \mathbf{M}[i,:] \cdot {\mathbf{\Phi}^{\left(i\right) \top}_{best}} \mathbf{z}^{\left(i\right)}$
\ENDFOR
  \ENDIF
\STATE Repeat the mask along mini-batch dimension:
    $\mathbf{M} \leftarrow \mathbf{M}.repeat(b, 1, 1)$
\STATE Apply the mask: $\bar{\mathbf{X}} = \mathbf{X} \cdot \mathbf{M}$
\STATE Normalize the features: \\
\# $sum(\mathbf{M})/$bcd is the empirical keep rate \\
$\bar{\mathbf{X}} \leftarrow \bar{\mathbf{X}} \cdot $bcd$ / sum(\mathbf{M})$ 
\STATE return $\bar{\mathbf{X}}$
\end{algorithmic}
\end{algorithm}

\section{Insights on DropCluster Optimization}\label{sec:objective}

In this section, we study the objective function of DropCluster from the perspective of \cite{DBLP:conf/cvpr/PalLVH20}, who analyzed Dropout and DropBlock applied to single hidden-layer linear networks, extending analysis on Dropout \cite{cavazza2018dropout,mianjy2018implicit}.
The analysis of \cite{DBLP:conf/cvpr/PalLVH20} can be extended to CNNs, by treating a convolutional layer as a special case of a fully-connected layer, with the kernel dimensions equal to the input dimensions. 
We consider training a single hidden-layer linear neural network with the squared loss after applying regular Dropout, which can be seen as optimizing the following stochastic objective:

\begin{equation}
       \min\limits_{\mathbf{U,V}}\mathbb{E}_\mathbf{z}\left\|\mathbf{Y-U}\text{diag}(\bm{\mu})^{-1}\text{diag}(\mathbf{z})\mathbf{V}^\top \mathbf{X}\right\|^{2}_{F}
\label{eq:opt_l2_dropout}
\end{equation}

\noindent Here, $\mathbf{X} \in \mathbb{R}^{b \times N}$ and $\mathbf{Y} \in \mathbb{R}^{a \times N}$ denote the features and labels of $N$ training samples, respectively. $\mathbf{U} \in \mathbb{R}^{a \times d}$ and $\mathbf{V} \in \mathbb{R}^{b \times d}$ are the output and input weights of the hidden layer with $d$ hidden features. 
The $d$-dimensional vector $\mathbf{z}$ contains the dropout variables of the 
neurons that are stochastically sampled at each training iteration with keep probability $\theta$. Let $z_i$ be the i$^{th}$ entry of $\mathbf{z}$, following the regular Dropout strategy, $z_i \sim \mathbf{Ber}(\theta)$, $\bm\mu$ is also a $d$-dimensional vector which is the mean of $\mathbf{z}$, i.e. $\mu_i = \mathbb{E}[z_i]$. Based on the results of \cite{cavazza2018dropout}, the optimization problem in Eq.~\ref{eq:opt_l2_dropout} is related to the following matrix approximation problem. Furthermore, we can express the approximations of Eq.~\ref{eq:opt_l2_dropout} in a deterministic formulation with a regularization term $\mathbf{\Omega}_{\mathbf{C}, \boldsymbol{\mu}}$ such that:
\begin{align}
    & \mathbb{E}_\mathbf{z}\left\|\mathbf{U}\text{diag}(\bm{\mu})^{-1}\text{diag}(\mathbf{z}) \mathbf{V}^\top \mathbf{X}\right\|^{2}_{F} \nonumber \\
    & = \left\|\mathbf{Y-U} \mathbf{V}^\top \mathbf{X}\right\|^{2}_{F} + \Omega_{\mathbf{C}, \boldsymbol{\mu}}(\mathbf{U}, \mathbf{X}^\top \mathbf{V}) 
    \label{eqn:deterministicr_form}
\end{align}
where $\mathbf{C}= \mbox{Cov}(\mathbf{z},\mathbf{z})$ is the covariance matrix of the dropout variables. The sampling strategy of $\mathbf{z}$ is determined by the type of dropout approach, and also changes the structure of 
$\mathbf{C}$. For instance, when $\mathbf{C}$ is diagonal, Eq.~\ref{eq:opt_l2_dropout} reduces to the objective of standard Dropout. 

We extend the analysis above to cover DropCluster. Specifically, 
the dropout variables are
not element-wise i.i.d but 
cluster-wise i.i.d. 
Let the $k$-dimensional vector $\mathbf{z'}$ contain the cluster dropout variables, $z'_i \sim \mathbf{Ber}(\theta)$ and $\mu'_i = \mathbb{E}[z'_i]$. $\boldsymbol{\Phi} \in \mathbb{R}^{k \times d}$ is the feature assignment matrix as defined in Sec.~\ref{sec:feature_assigment}. 
By introducing $\mathbf{z}=\Phi^{\top}\mathbf{z'}$ and $\bm{\mu}=\Phi^{\top}\bm{\mu'}$, the objective function of DropCluster can be written as

\begin{equation}
       \min\limits_{\mathbf{U,V}}\mathbb{E}_\mathbf{z}\left\|\mathbf{Y-U}\text{diag}(\bm{{\Phi^\top}\mu'})^{-1} \text{diag}(\mathbf{{\Phi^\top}z'})\mathbf{V}^\top \mathbf{X}\right\|^{2}_{F}
\label{eq:opt_l2_dropcluster}
\end{equation}

Eq.~\ref{eqn:deterministicr_form} also holds for Eq.~\ref{eq:opt_l2_dropcluster}. We further derive the regularization term from the analysis of Pal et al.~\cite{DBLP:conf/cvpr/PalLVH20} (see Eqs. 17 and 18) on structured Dropout to get a particular form of $\mathbf{\Omega}_{\mathbf{C}, \boldsymbol{\mu}}$ for DropCluster.

\begin{align}
 \Omega_{\mathbf{C}, \boldsymbol{\mu}}(\mathbf{U}, \mathbf{X}^\top \mathbf{V})
        & = \langle \bar{\mathbf{C}}, \mathbf{U}^\top \mathbf{U} \odot  \mathbf{V}^\top \mathbf{X} \mathbf{X}^\top \mathbf{V} \rangle  \notag\\
   & = \sum_{i, j = 1}^d \frac{c_{i, j}}{\mu_i \mu_j} (\mathbf{u}_i^\top \mathbf{u}_j)(\mathbf{h}_i^\top\mathbf{h}_j) \notag \\
   & = \sum_{g=1}^{k} \sum_{i, j \in g} \mbox{constant} \cdot (\mathbf{u}_i^\top \mathbf{u}_j)(\mathbf{h}_i^\top\mathbf{h}_j)
   \label{eqn:regularizer_term}
\end{align}

where $\mathbf{h}_i$ denotes the $i$-th column of the output of the hidden layer, i.e. $\mathbf{X}^\top \mathbf{V}$, and $\bar{\mathbf{C}}$ is defined as the characteristic matrix in terms of the mean and covariance of the per-neuron dropout variables $\mathbf{z}$,
\begin{align}
    \bar{c}_{i,j}&=\frac{c_{i,j}}{\mu_i \mu}_j &\text{or} &&\bar{\mathbf{C}}= \text{diag}(\bm{\mu})^{-1}\mathbf{C}\text{diag}(\bm{\mu})^{-1}
\end{align}

Similar to original Dropout \cite{JMLR:v15:srivastava14a}, dropout-like methods induce regularization by averaging the predictions of multiple possible settings of a fixed-sized model. Thus, $\bar{\mathbf{C}}$ can be interpreted as a re-scaling of the parameters of the model to average up to the approximations in Eq.~\ref{eqn:deterministicr_form}. 
In standard Dropout, $\mathbf{C}$ is diagonal with entries $c_{i,j}=\theta(1-\theta)$, hence $\bar{\mathbf{C}}$ is also diagonal with entries $\bar{c}_{i,j}=\frac{(1-\theta)}{\theta}$. 
In DropBlock with block-size $r$, ${z_i}$ is block-wise i.i.d. Thus, $\bar{\mathbf{C}}$ is a \textit{block} diagonal matrix as $\bar{\mathbf{C}}=\frac{(1-\theta)}{\theta}\mbox{BlkDiag}(\mathbf{1}_r, \mathbf{1}_r^\top, \cdots, \mathbf{1}_r, \mathbf{1}_r^\top)$, here $\mbox{BlkDiag}(\cdot)$ is a matrix formed by $r$-size square blocks lying on the diagonal with all ones. In the case of DropCluster, the structure of $\bar{\mathbf{C}}$ is similar to DropBlock, but with blocks that do not necessarily contain an equal number of features due to the arbitrary shape of the clusters. 
For DropCluster with $k$ clusters, $\bar{\mathbf{C}}$ has the following form,
\begin{align}
    \bar{\mathbf{C}} &= \frac{(1-\theta)}{\theta}\mathbf{\Phi}^\top\mathbf{\Phi} 
    =\frac{(1-\theta)}{\theta}\mbox{BlkDiag}(\mathbf{1}_{r_1} \mathbf{1}_{r_1}^\top, \cdots, \mathbf{1}_{r_k} \mathbf{1}_{r_k}^\top)
    \label{eqn:characteristic_matrix}
\end{align} 
where $\sum_{i=1}^{k}r_i = d$. 
In summary, DropCluster shares the properties described by \cite{DBLP:conf/cvpr/PalLVH20,cavazza2018dropout,mianjy2018implicit} with respect to convergence to a minimum of the objective function and to the properties of this minimum due to the regularization it induces. Similar to Dropout and DropBlock, the weights of the network are balanced, but across clusters instead of neurons or blocks. Additionally, similar to DropBlock, DropCluster also induces spectral k-support norm resulting in sparser solutions compared to Dropout. (This analysis only holds precisely for the case of a single hidden layer network.)

%% file: sec/Experiments.tex
In this section, we empirically evaluate the effectiveness of DropCluster on benchmark datasets. 
Similar to Zunino et al.~\cite{DBLP:journals/ijcv/ZuninoBMZSM21}, we also use light networks and small datasets because our aim is to study overfitting and few-shot learning in real-world applications.
Specifically, we train LeNet-5 \cite{lecun1989backpropagation} and AlexNet \cite{krizhevsky2012imagenet} on the CIFAR-10/100 \cite{krizhevsky2009learning}, the Street View House Numbers (SVHN) \cite{netzer2011reading} and the APPA-REAL \cite{clapes2018apparent} datasets with three \textbf{dropout types}: Dropout, DropBlock and DropCluster, and compare their performance in terms of loss values. 
We also use a network without dropout, dubbed \textit{No-drop}, as a baseline. 

\begin{figure}[b]
	\centering
	\includegraphics[width=0.6\textwidth]{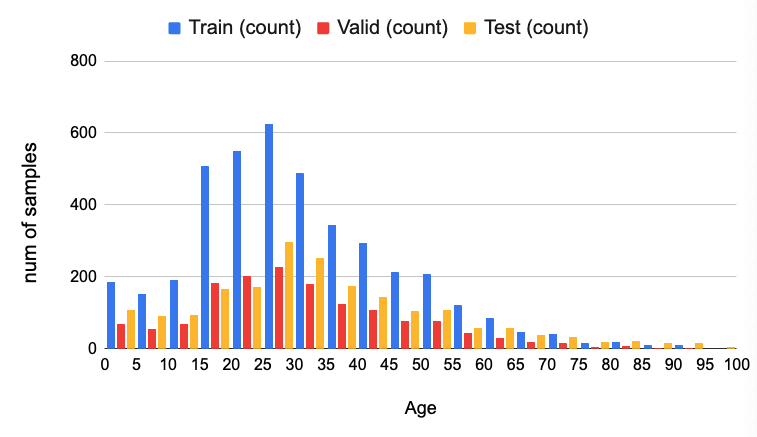}
	\caption{Histogram of real age labels for APPA-REAL dataset.}
	\label{fig:age_label}
\end{figure}

\subsection{Datasets} 
In this paper, we conduct experiments on three publicly available image datasets. Details for each dataset
such as image resolution and pre-processing are summerized below.

\textbf{CIFAR-10/100} are a widely used datasets that consist of  $32 \times 32$ natural RGB images from 10/100 classes respectively, with 600 images per class \cite{krizhevsky2009learning}.  
The standard split contains 500 training images and 100 test images per class. We use $80\%$ of the training set for training, the remaining $20\%$ as a validation set to decide if overfitting occurs, and the full test data for evaluation.

\textbf{Street View House Numbers (SVHN)}~\cite{netzer2011reading} is a digit image dataset captured from Google Street View. The dataset is composed of $73,257$ training images and $26,032$ test images. Each image has been resized to a fixed $32 \times 32$ resolution with a single digit from $0$-$9$ in the center. Note that the categories are not balanced. Similar to CIFAR-10/100, we split $20\%$ of the training set for validation.

\textbf{APPA-REAL}~\cite{clapes2018apparent} is a dataset of face images from subjects whose age spans $0$ 
to $100$ years old. It comes with real and apparent age labels; we only use the former in our experiments. Unlike CIFAR-10/100, where the distribution of samples from different categories is uniform, this is not the case for APPA-REAL. Figure \ref{fig:age_label} shows the distribution of real age labels in the APPA-REAL dataset. The dataset is much smaller than CIFAR-10 and SVHN, only containing $4,113$ training, $1,500$ validation and $1,978$ test images. Since the age groups are not balanced, samples are rare at the tails of the distribution. We merge the age labels in 5-year groups to obtain a total of 20 categories in the classification task on this dataset.

\textbf{Pre-pocessing step} for CIFAR-10/100 and SVNM only includes normalization for each sample. Since all photos in APPA-REAL are taken from different cameras with different resolutions, pre-processing includes more steps: we first apply center-crop and resize the images to $64 \times 64$ pixels, then apply normalization. 

\subsection{Metrics} 
We choose loss over classification accuracy as the criterion because overfitting is detected by observing the training and validation losses and because achieving lower loss is an indication of robustness for a network. Classification or regression accuracy are byproducts of trained networks, and thus provide indirect evidence for overfitting. We also pay attention to the gap between validation and training loss which is an indicator of generalization. Visualization of training/validation loss curves is available in Fig.\ref{fig:experiment_results}, while Table\ref{tab:exp_result} provides average training/test losses and standard deviations of test loss in the last epoch. In particular, the best and second best scores are highlighted in red and yellow, respectively, throughout the tables.

\begin{table}[htb]
\caption{Comparison of dropout types on LeNet-5 and AlexNet (only when noted). We compute the Cross-Entropy loss for the four classification tasks on the left, and the $L_1$ loss for the rightmost regression task. We report average loss values and corresponding standard deviations after 10 runs.
	}
\label{tab:exp_result}
\vskip 0.15in
\centering
\adjustbox{width=0.9\textwidth}{
	\begin{tabular}[width=\columnwidth]{lccccccccc}
		\toprule
		& \multicolumn{3}{c}{CIFAR-10} 
		& \multicolumn{3}{c}{CIFAR-100} 
            & \multicolumn{3}{c}{CIFAR-100-AlexNet} \\
		\cmidrule(r){2-10}
		 & \textbf{Train}  &\textbf{Test} & \textbf{Std} 
		 & \textbf{Train}  &\textbf{Test} & \textbf{Std}
          & \textbf{Train}  &\textbf{Test} & \textbf{Std} \\
		
		\midrule
		No-drop & 0.865	& 1.084	& \colorbox{red!25}{0.014}
		        &1.614	& 3.293 & 0.152  
                    &0.308	& 5.020	& 0.633  \\
		Dropout &  0.647 &\colorbox{yellow!25}{0.980} &0.034	
		        &2.163	&2.737	&0.036 
                    &1.388	&\colorbox{yellow!25}{2.347}	&\colorbox{yellow!25}{0.036} \\
		DropBlock & 0.686 &1.040 &0.080	
		         &2.399	&2.716	& \colorbox{red!25}{0.028}	
                   &1.638	&2.419	&0.048    \\
		DropCluster & 0.868	&\colorbox{red!25}{0.915}	&\colorbox{yellow!25}{0.017}
		            &2.276	&\colorbox{red!25}{2.657}	&\colorbox{yellow!25}{0.029}	
                        &1.983	&\colorbox{red!25}{2.228}	& \colorbox{red!25}{0.016}  \\		           
        \midrule
            & \multicolumn{3}{c}{SVHN}
		&  \multicolumn{3}{c}{APPA-REAL-CLF} 
		& \multicolumn{3}{c}{APPA-REAL-REG} \\
		\cmidrule(r){2-10}
            & \textbf{Train}  &\textbf{Test} & \textbf{Std}
		  & \textbf{Train}  &\textbf{Test} & \textbf{Std}
		  & \textbf{Train}  &\textbf{Test} & \textbf{Std} \\
		  \midrule
		No-drop   &0.187	&0.480	&0.026
                     &0.104	&12.743	&1.184	
		        &0.305	&10.200	&\colorbox{yellow!25}{0.158}  \\
		Dropout  &0.335	&0.385	&0.016
                    &1.255	&\colorbox{yellow!25}{2.516}	&\colorbox{yellow!25}{0.119}	
		        &1.397	&\colorbox{yellow!25}{10.025}	&0.169 \\
		DropBlock &0.379	&\colorbox{yellow!25}{0.383}	&\colorbox{yellow!25}{0.013}
                  &1.357	&2.597	&0.129	
		         &1.867	&10.283	&0.172 \\
		DropCluster &0.339	& \colorbox{red!25}{0.358} &\colorbox{red!25}{0.007}
                        &1.813	&\colorbox{red!25}{2.346}	&\colorbox{red!25}{0.025}	
		            &2.063	&\colorbox{red!25}{9.976}	&\colorbox{red!25}{0.154} \\
		 \bottomrule
	\end{tabular}
 }
\end{table}

\subsection{Implementation Details} 

\subsubsection{Experimental Setup}
We adopt LeNet-5 and AlexNet architectures, introduced a dropout layer after the initial convolutional and pooling layers. All networks consistently utilized a batch size of $64$. The determination of the drop probability parameters for each dropout type is elucidated in the ablation study detailed in Appendix \ref{sec:ablation_study}. Each experiment is repeated across 10 different random initializations, with the best dropout probability for each dropout type.

The block size of $5 \times 5$ is employed in DropBlock experiments, aligning with the publicly available implementation of DropBlock \cite{dropblock_github}. As described in Section~\ref{sec:dropcluster}, DropCluster were applied after the model exhibits signs of overfitting, considering the variable nature of structure information at the beginning of training. Practical implementation involves applying k-means \cite{lloyd1982least} to derive the feature assignment matrix from a mini-batch of 300 samples uniformly selected from the training set across all categories. Although the clustering process, inclusive of Silhouette Coefficient computation, incurs a time cost of 61.31 s on CIFAR-100, the subsequent training iterations maintain a competitive speed compared to alternative methods. For example, training LeNet5 on CIFAR-100 with DropCluster takes 31.93 s per epoch, showcasing minimal slowdowns of 6.55\%, 5.72\%, and 2.76\% compared to No-drop, Dropout, and DropBlock, respectively. We also conducted an exploration of DropCluster application across different layers in Section~\ref{sec:drop_over_layers}. Further intricacies related to network architecture modifications are extensively explained in Appendix \ref{sec:architecture}.

\subsubsection{Training Strategy, Software and Hardware Platforms}
All classifiers are trained using Cross-Entropy loss \cite{murphy2012machine}, employing SGD with the Adam optimizer \cite{kingma2014adam} and a weight decay parameter set to $1e-4$. A scheduler with an initial learning rate of $0.001$ is utilized, incorporating a multiplicative learning rate decay of $0.1$ at the $50^{th}$ and $100^{th}$ epoch. All experiments are implemented using Python 3.7 \cite{oliphant2007python}, PyTorch \cite{paszke2019pytorch}, scikit-learn \cite{pedregosa2011scikit}, and NumPy \cite{walt2011numpy}, executed on a Nvidia TITAN RTX GPU with 24 GB memory and a 24-core Intel CPU.

\subsection{Quantitative Results}

\subsubsection{CIFAR-10/100 Image Classification}
For both CIFAR-10 and 100 \cite{krizhevsky2009learning}, we use $80\%$ of the training set to train LeNet-5 with the three dropout types, the remaining $20\%$ as a validation set to detect overfitting, and the full test data for evaluation. 

According to Table~\ref{tab:exp_result} and Fig.~\ref{fig:experiment_results} (a) and (b), DropCluster achieves the lowest test loss and the smallest generalization gap compared to the baselines. 

We also report the training/test accuracy and the standard deviation of test accuracy of LeNet-5 trained on the CIFAR-10 dataset in Table \ref{tab:exp_accuracy}. %
Since our training is under overfitting conditions, the reported accuracy is not as high as that of well-trained and well-fitting models. DropCluster achieved better results in test accuracy and smaller standard deviation than other dropout methods, which is consistent with the observed loss.

We also conduct experiments with AlexNet on the CIFAR-100 dataset (Table~\ref{tab:exp_result} and Fig.~\ref{fig:experiment_results} (c)). Significant overfitting occurs when training the No-drop model on CIFAR-100, but it is prevented using all regularizers. Fig.~\ref{fig:experiment_results} (c) shows how the solid blue curve (validation) closely tracks the dashed blue curve (training) for DropCluster. 

\begin{table}[hb]
\small
\centering
\caption{Classification Accuracy of LeNet-5 train on CIFAR-10 datasets with different dropout types.}
\vskip 0.15in
\adjustbox{width=0.7\textwidth}{
\label{tab:exp_accuracy}
	\begin{tabular}{lcccccc}
		\toprule
		& \multicolumn{3}{c}{\textbf{Acc@1}} 
            & \multicolumn{3}{c}{\textbf{Acc@5}} \\
		\cmidrule(r){2-7}
		 & \textbf{Train}
            & \textbf{Test} 
		 & \textbf{Std} 
           & \textbf{Train}
            & \textbf{Test} 
		 & \textbf{Std}  \\
		\midrule
		No-drop & 99.816 & 60.891	
		        & \colorbox{red!25}{0.461} 
                    & 99.997 
                    & 94.970	
		        & \colorbox{yellow!25}{0.221}\\
		Dropout &  76.477 
                    & \colorbox{yellow!25}{67.389}	
		        & 0.954	 
                    & 98.783 
                    & \colorbox{yellow!25}{96.776}
		        & 0.290\\
		DropBlock & 75.748 & 66.229	
		          & 1.135	
                    & 98.634 & 96.437	
		          & 0.387\\
		DropCluster & 69.239 
                        & \colorbox{red!25}{68.213}	
		            & \colorbox{yellow!25}{0.530} 
                        & 97.637 
                        & \colorbox{red!25}{97.123}	
		              & \colorbox{red!25}{0.190}\\	
		 \bottomrule
	\end{tabular}
 }
\end{table}

\subsubsection{SVHN Digit Classification}

Because the SVHN dataset \cite{netzer2011reading} is small, we only use the LeNet-5 architecture here. 
The three dropout types perform similarly to the CIFAR-10 experiment due to similarities between CIFAR-10 and SVHN, as shown in Fig.~\ref{fig:experiment_results} (d). Although the three dropout types yield similar results, the training and validation losses of DropCluster represented by the blue curves are almost overlapping, while it achieves the lowest validation loss.

\newcommand\widf{.33\textwidth}

\begin{figure*}[tb]
\begin{center}
\adjustbox{width=\textwidth}{
\begin{tabular}{ccc}
\includegraphics[width=\widf]{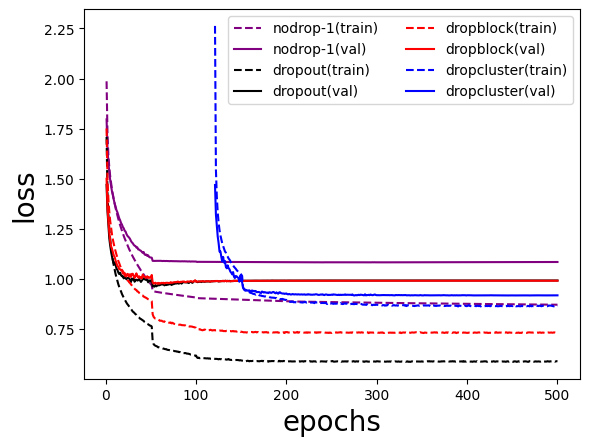} &
\includegraphics[width=\widf]{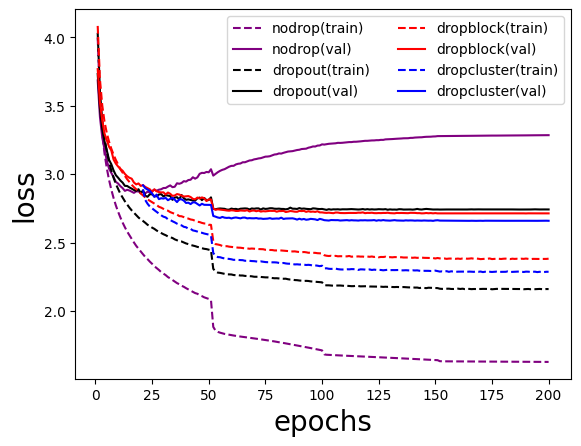} & 
\includegraphics[width=\widf]{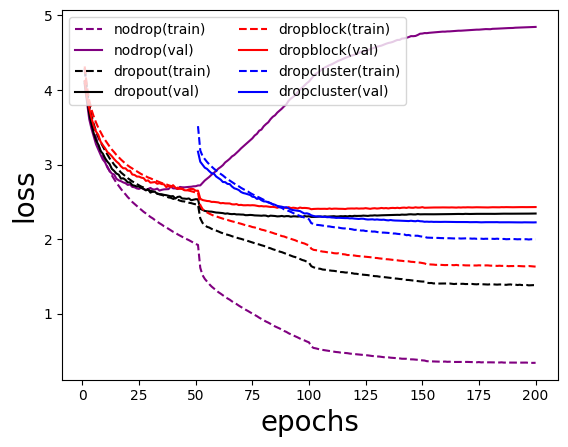} \\
(a) LeNet-5 on CIFAR-10 &
(b) LeNet-5 on CIFAR-100 &
(c) AlexNet on CIFAR-100 \\

\includegraphics[width=\widf]{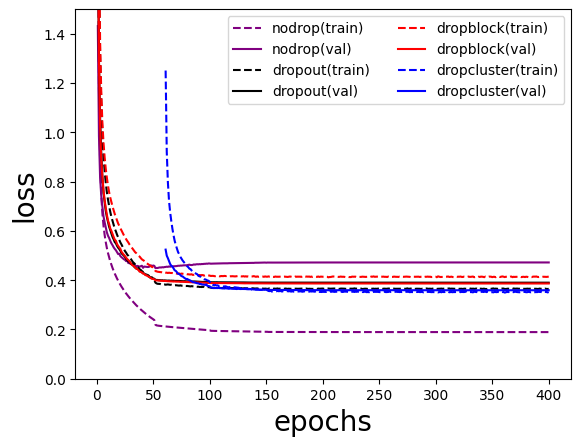} &
\includegraphics[width=\widf]{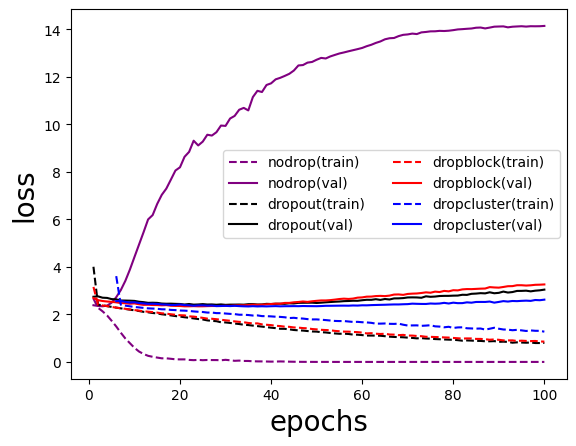} &
\includegraphics[width=\widf]{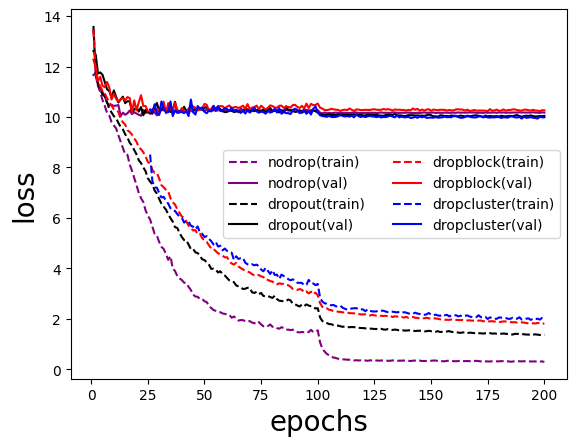}\\
(d) LeNet-5 on SVHN &
(e) LeNet-5 on APPA-REAL-CLF &
(f) LeNet-5 on APPA-REAL-REG \\
\end{tabular}
}
\end{center}
\vspace{-4pt}
\caption{Training/Validation loss for different dropout types. The purple curves represent the No-drop model that begins overfitting after several epochs. The red solid curves show that DropCluster models reach better performance than the other two approaches in CIFAR-10/100 and SVHN experiments. The classification task on APPA-REAL data is especially challenging for No-drop, while DropCluster controls overfitting. The smaller difference between validation and training loss of DropCluster indicates better less overfitting and generalization.}
\label{fig:experiment_results}
\end{figure*}

\subsubsection{APPA-REAL Age Estimation}
APPA-REAL \cite{clapes2018apparent} is a dataset of face images from subjects whose age spans $0$ to $100$ years old. Age estimation on this dataset is a typical few-shot learning problem, since images of elderly and young people are limited. (See Fig. \ref{fig:age_label}.)
We formulate age prediction from a single image 
as a classification or as a regression problem.

In \textit{classification}, the goal is to assign each query image to the corresponding age group. Since the age groups are not balanced, samples are rare at the tails of the distribution which makes it difficult for the models to learn from only one or two images. Thus, we merge the age labels in 5-year groups to obtain 20 categories in total, and train LeNet-5 with Cross-Entropy Loss.

In \textit{regression}, the goal is to predict the real age given a face image. We add a $3$-layer MLP at the end of the LeNet-5 architecture, as shown in Fig. \ref{fig:architecture} (b) in Appendix \ref{sec:architecture}, to make the final prediction, and train it using the $L_1$ loss. 

In the classification task, all three methods decrease the validation loss by more than $50\%$ (Fig.~\ref{fig:experiment_results} (e)). Although they still overfit at the later epochs, Dropout and DropCluster outperform DropBlock, and DropCluster generates an almost flat curve. In the regression task, all algorithms overfit. DropBlock overlaps with No-drop, while Dropout and DropCluster have lower validation losses (Fig.~\ref{fig:experiment_results} (f)). The gap between the training and validation curves is smaller for DropCluster than the alternatives.

\subsection{DropCluster on Different Layers}
\label{sec:drop_over_layers}

\begin{figure}[h]
	\centering
	\includegraphics[width=0.7\textwidth]{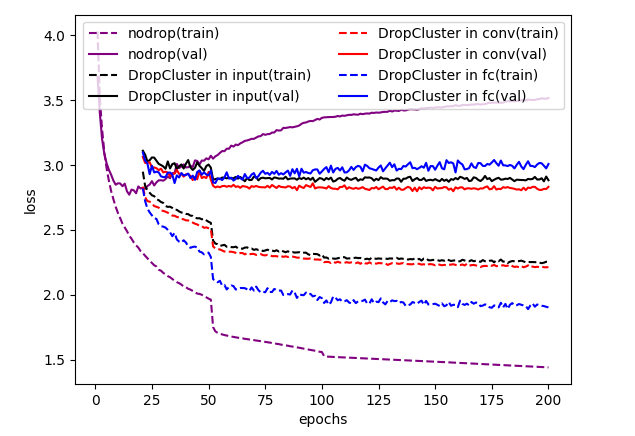}
	\caption{
  Training and validation losses for DropCluster on image clusters (input), a convolutional (conv) and a fully connected (fc) layer of LeNet-5 on CIFAR-100. }
	\label{fig:dropcluster_diff_layer}
\end{figure}

We present an experiment on LeNet-5 trained on CIFAR-100 with DropCluster on different layers, see Fig.~\ref{fig:dropcluster_diff_layer}. Specifically, we perform DropCluster on clusters obtained in the input images, on convolutional and fully connected layers. Performing clustering 
on the RGB image can be seen as a weaker form of clustering than applying k-means in feature space. It can be seen in the figure that clustering in feature space (‘‘DropCluster in conv’’) is somewhat better, as expected.

We also test DropCluster on a fully connected layer near the output. Due to the preceding pooling and activation layers, the features of this layer contain global statistical information and are loosely attached to the spatial properties of the input limiting the benefits of dropout-style methods. \cite{JMLR:v15:srivastava14a,tompson2015efficient} suggest that dropout should be applied in the earlier layers that have stronger spatial correlations, consistently with Fig.~\ref{fig:dropcluster_diff_layer} (red vs. blue).

\subsection{Discussion} 
CNNs are likely to overfit when trained on small datasets. Dropout, DropBlock and DropCluster prevent overfitting in the CIFAR-10/100 and SVHN experiments. 
Even in the APPA-REAL classification experiment, where the much smaller dataset causes severe overfitting, the three dropout regularizers 
are effective. This is not the case in the regression experiment though.
Comparing the results in Table~\ref{tab:exp_result}, DropCluster outperforms the other approaches in all classification experiments, and has the smallest standard deviation and generalization gap in most of them.

DropCluster's limitation is that it requires features to exhibit clustering behavior. Clusters, however, may become less prominent after multiple pooling and convolution operations.
This limits DropCluster's effectiveness on layers far from the input. 
DropCluster should be extended to be applicable on networks where correlated data can be propagated along multiple pathways, such as those with skip connections. To effectively drop information that is propagated across layers via a convolutional path and a skip connection, both paths would have to be dropped simultaneously. This would require the development of a mechanism for detecting correlations across paths of different types.

%% file: sec/Conclusion.tex
We propose DropCluster, a novel data-driven regularizer for CNNs. It is based on identifying correlated information in the data by clustering the features of a network layer and applying dropout on the clusters. We also propose the Spatial Hopkins statistic to assess clustering tendency and show that features in convolutional layers form clusters. Our experimental results 
show the effectiveness of DropCluster in a variety of settings. 
As shown in Table~\ref{tab:exp_result}, DropCluster achieves the lowest test loss in all six experiments, and the lowest standard deviation and generalization gap in most of them.
Our findings indicate that leveraging structure in convolutional layers when training CNNs can prevent overfitting more effectively than regular Dropout and DropBlock, which do not adapt to the data.

%% file: sec/Supplement.tex
\setcounter{figure}{0}
\setcounter{table}{0}
\renewcommand\thefigure{A.\arabic{figure}}  
\renewcommand\thetable{A.\arabic{table}}
\renewcommand\thesubsection{A.\arabic{subsection}}
\setcounter{algorithm}{1}

This appendix presents the algorithm for discovering the number of clusters (Section~\ref{sec:compute_cluster});   the modifications to the standard architectures to enable DropCluster (Section~\ref{sec:architecture}); and an ablation study on the selection of drop probability (Section~\ref{sec:ablation_study}).

\subsection{Discovering the number of Clusters}
\label{sec:compute_cluster}

Algorithm \ref{alg:compute_clusters} shows the procedure for discovering the number of clusters based on the Silhouette Coefficient and computing the feature assignment matrix, which is described in Section \ref{sec:dropcluster} of the main paper. 
This algorithm should be run before applying dropout on the clusters and is applied only \textit{once} during the entire training process. Any clustering algorithm can be applied to compute clusters in line 4 of Algorithm~\ref{alg:compute_clusters}; we use k-means in our experiments.

\begin{algorithm}[H]
  \caption{Compute Clusters}
  \textbf{Input:} 

  \hspace{\parindent} $\mathbf{X} \in \mathbb{R}^{b \times c \times d}$, output activations of the convolutional layer, $b$ is the mini-batch size, $c$ is the number of channels and $d$ is the size of the feature map at each channel of the given layer.

  \hspace{\parindent} $[m,n]$: search range for number of clusters.
   

  \textbf{Output:} 

  \hspace{\parindent} Set of $\mathbf{\Phi}_{best}^{(i)} \in \{0,1\}^{k \times d}$, best feature assignments for channel $i$

  \label{alg:compute_clusters}
\begin{algorithmic}[1]
  \FOR{$i=1$ {\bfseries to} $c$}
  \STATE Initialize best Silhouette Coefficient: $s_{best} = -1$
  \FOR{$j=m$ {\bfseries to} $n$}
  \STATE Let $\mathbf{\Phi}_j^{(i)} = compute\_clusters(\mathbf{X[:, i, :]}, j)$, $\mathbf{\Phi}_{j}^{(i)} \in \{0,1\}^{j \times d}$ 
  \STATE Computer Silhouette Coefficient $s_j$ for $\mathbf{\Phi}_j^{(i)}$
  \IF{$s_j > s_{best}$} 
  \STATE $\mathbf{\Phi}_{best}^{(i)}$ = $\mathbf{\Phi}_j^{(i)}$ 
  \ENDIF
  \ENDFOR
  \ENDFOR
  \STATE Return {$\{\mathbf{\Phi}_{best}^{(i)}$: for $i \in [1,c]\}$}
\end{algorithmic}
\end{algorithm}

\subsection{Network Architecture Modifications}
\label{sec:architecture}
\begin{figure}[thb]
\begin{center}
\begin{tabular}{cc}
\includegraphics[width=0.5\textwidth]{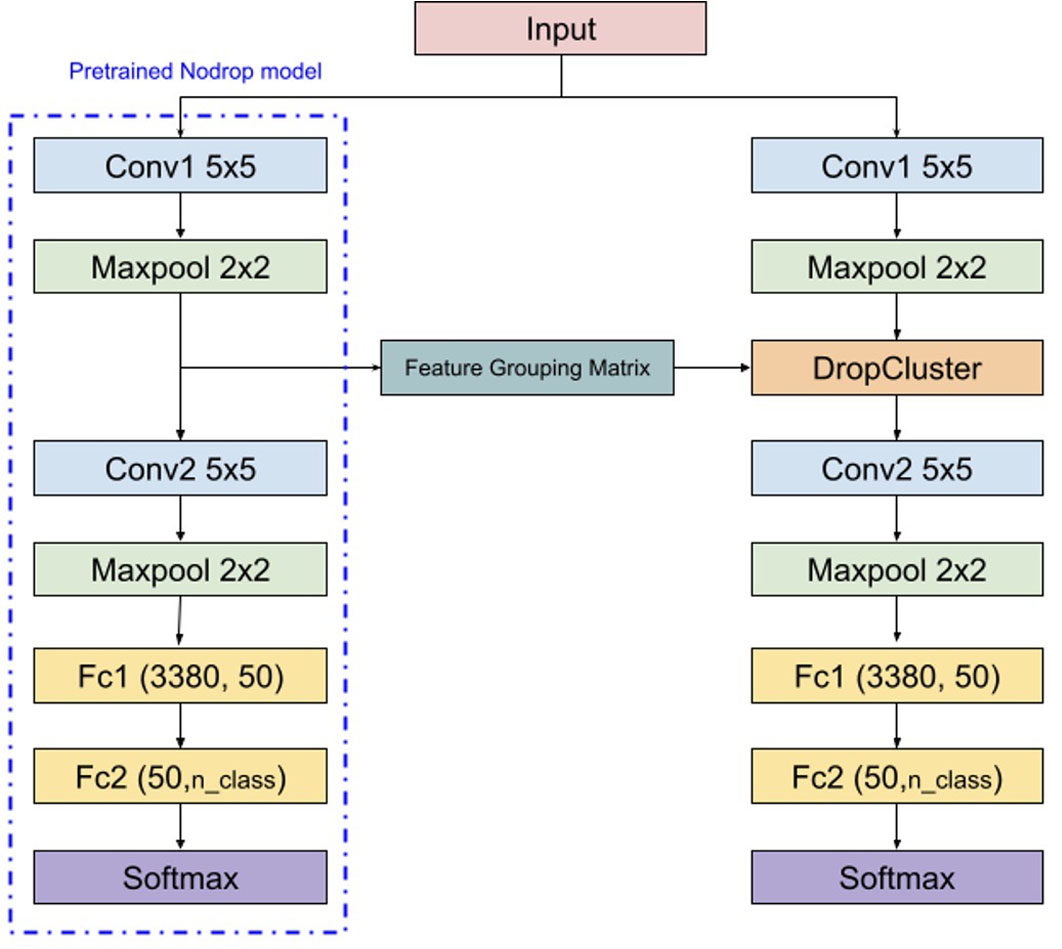} 
& \includegraphics[width=0.5\textwidth]{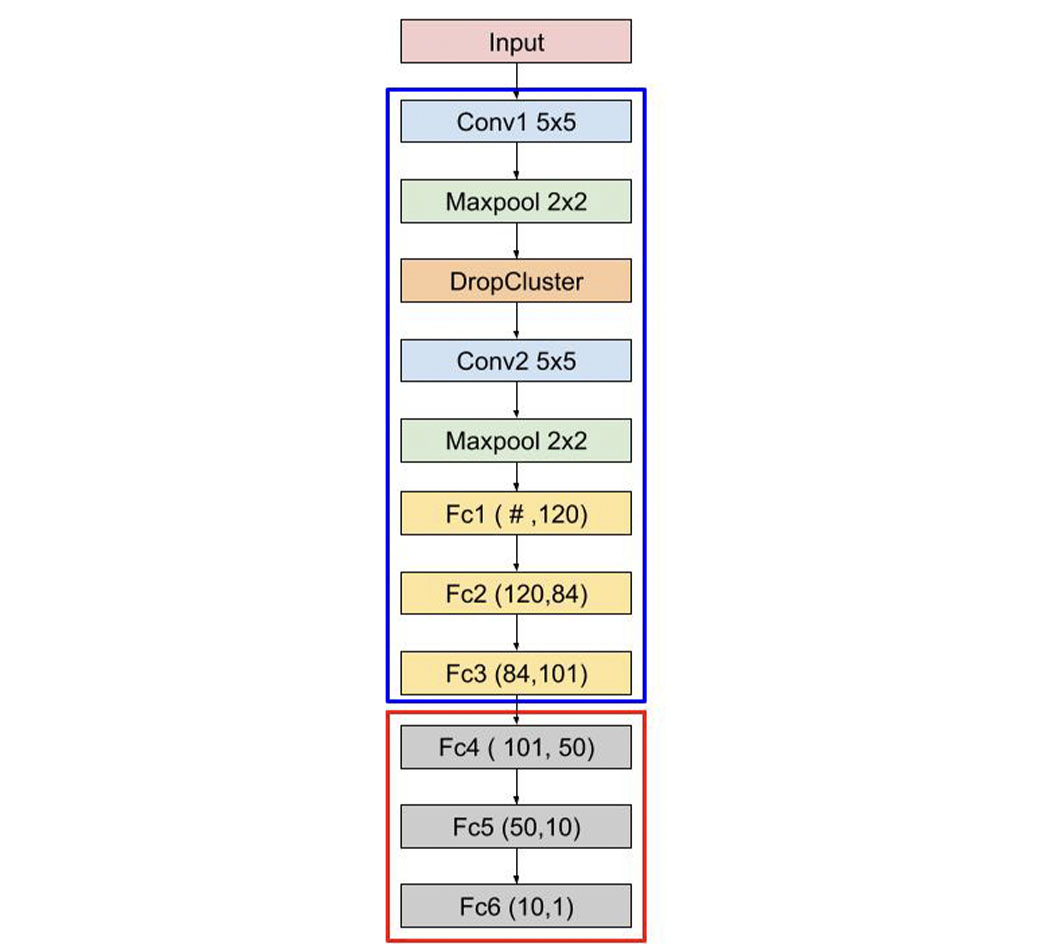}\\
(a) Inserting DropCluster layer in LeNet-5  
& (b) LeNet-5 modification for regression\\
\end{tabular}
\end{center}
\caption{(a) The DropCluster layer is inserted after the first convolutional and pooling layer, and is activated after pre-training. (b) A 3-layer MLP is added at the end of LeNet-5 classification model to compute the final prediction. Other dropout layer types are inserted similarly to the DropCluster layer but not starting from the pre-trained model.}
\label{fig:architecture}
\end{figure}

In this section, we discuss how we modify the standard LeNet-5 and AlexNet architectures to enable Dropout, DropBlock and DropCluster for classification and regression tasks. The three dropout regularizers are implemented by inserting the corresponding dropout layer.
Figure~\ref{fig:architecture} (a) shows how to apply DropCluster to LeNet-5 for a classification problem: 
We first train LeNet-5 without any dropout until it begins overfitting. We save the checkpoint and the output activation values of the 
previous layer, and apply a clustering algorithm such as k-means to obtain the feature assignment matrix. We use this feature assignment matrix to set up the DropCluster layer and insert it into the saved pre-trained model, then continue the training from the checkpoint. For DropBlock and Dropout, we do not need to run the pre-training step. We just replace the ``DropCluster'' layer 
in the right column of Fig.~\ref{fig:architecture} (a) 
with a ``Dropout'' or ``DropBlock'' layer and start to train from the beginning. The modification to the AlexNet architecture is 
similar.

In the age estimation experiment, we train LeNet-5 for both classification and regression. The outputs of the two tasks are quite different: regression predicts one number, while classification predicts class probabilities. Hence, we modify the LeNet-5 architecture by adding a 3-layer MLP at the end of the classification model to compute the final prediction for regression as shown in Fig.~\ref{fig:architecture} (b).

In all experiments, we inserted the dropout layers after the first convolutional and pooling layer, because the size of the channels becomes too small to exhibit structure information in the subsequent layers. Ablation study on the dropout rate for all dropout types across different datasets is conducted in the next sections.

L. Figure \ref{fig:age_label} shows the distribution of real age labels in the APPA-REAL dataset. The dataset is much smaller than CIFAR-10 and SVHN, only containing $4,113$ training, $1,500$ validation and $1,978$ test images. Since all photos are taken from different cameras with different resolutions,  pre-processing includes more steps: we first apply center-crop and resize the images to $64 \times 64$ pixels, then apply normalization.

\subsection{Ablation Study on Drop Probability}
\label{sec:ablation_study}

In this section, we present an ablation study on the drop probability parameter across all datasets and dropout types. We vary the dropout probability $1-\theta$ for all dropout types from $0.1$ to $0.7$ with a step of $0.1$. We run each setting with three different random seeds, and record the average training/validation loss of each epoch. Fig. \ref{fig:ablation_cifar} and \ref{fig:ablation_svnh_age} show loss curves for each method for different values of $1-\theta$.
We also report the drop probability selection, based on the best validation loss, for each experiment in Tab. \ref{tab:drop_prob_list}. In all figures, the purple lines represent the No-drop model, which serves as the baseline. Notice that the drop probabilities for DropBlock in the table and figures have been adjusted to account for overlaps among blocks and correspond to the effective probability of a feature being dropped.

\begin{table}[htb]
\small
    \centering
    \caption{Drop probablity selection for each experiment}
    \vskip 0.15in
\adjustbox{width=0.7\textwidth}{
    \label{tab:drop_prob_list}
    \begin{tabular}{cccc}
    	\toprule
    	& \textbf{Dropout} & \textbf{DropBlock} & \textbf{DropCluster} \\
    	\midrule
    	CIFAR-10-LeNet-5 & 0.3 & 0.1 & 0.4 \\
    	CIFAR-100-LeNet-5 & 0.3 & 0.1 & 0.2 \\
    	CIFAR-100-AlexNet  & 0.5 & 0.1 & 0.6 \\
    	SVHN-LeNet-5 & 0.5 & 0.2 & 0.4\\
    	APPA-REAL-CLF-LeNet-5 & 0.7 & 0.6 & 0.7\\
    	APPA-REAL-REG-LeNet-5 & 0.1 & 0.1 & 0.1\\
        \bottomrule
    \end{tabular}
    }
\end{table}

In summary, the three dropout regularizers prevent overfitting in most of the experiments. Dropout is able to prevent overfitting in most of the classification experiments but has larger generalization gaps between training and validation loss than DropBlock and DropCluster. DropBlock is most sensitive to the drop probability, meanwhile the effective drop probability for DropBlock is limited to a small range. As the drop probability increases, blocks are more likely to overlap with each other, causing a smaller number of features to be dropped than expected. DropCluster is less sensitive to the drop probability than other alternatives, and achieves lower validation loss and smaller generalization gaps in all experiments.

\newcommand\awidf{.3\columnwidth}


\renewcommand{\tablename}{Figure}

%
%

\begin{table*}[h]
\centering
\begin{tabular}{ccc}
   \includegraphics[width=\awidf]{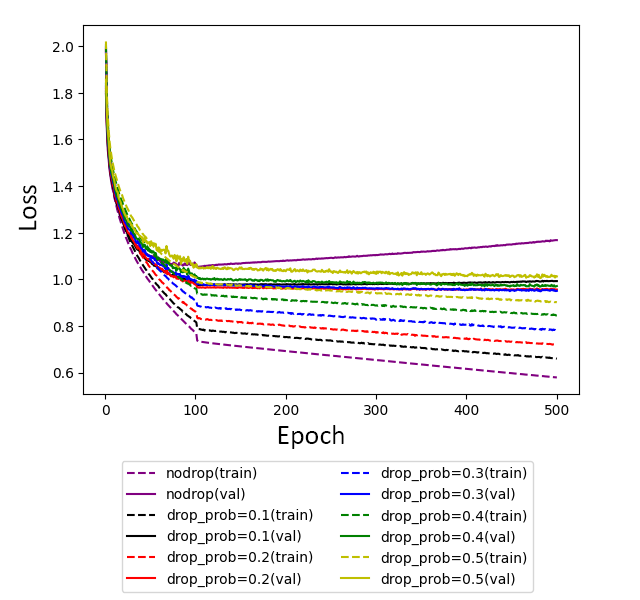} &
   \includegraphics[width=\awidf]{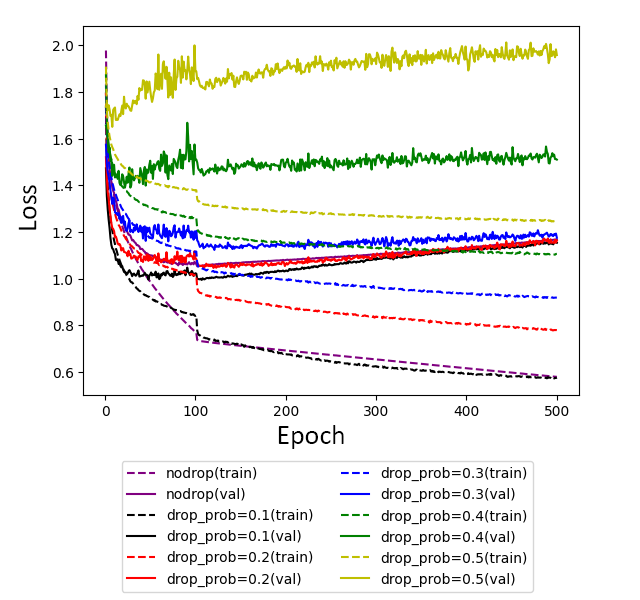} &
   \includegraphics[width=\awidf]{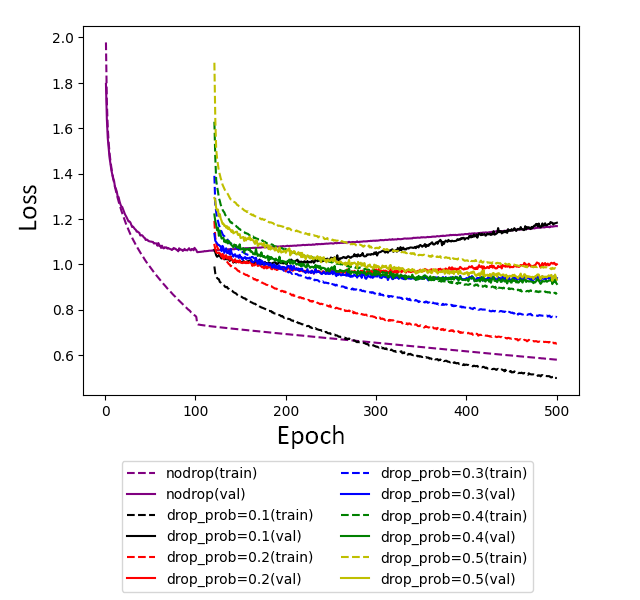} \\
(a) Dropout-CIFAR10-LeNet5 &
(b) DropBlock-CIFAR10-LeNet5 &
(c) DropCluster-CIFAR10-LeNet5 \\


\includegraphics[width=\awidf]{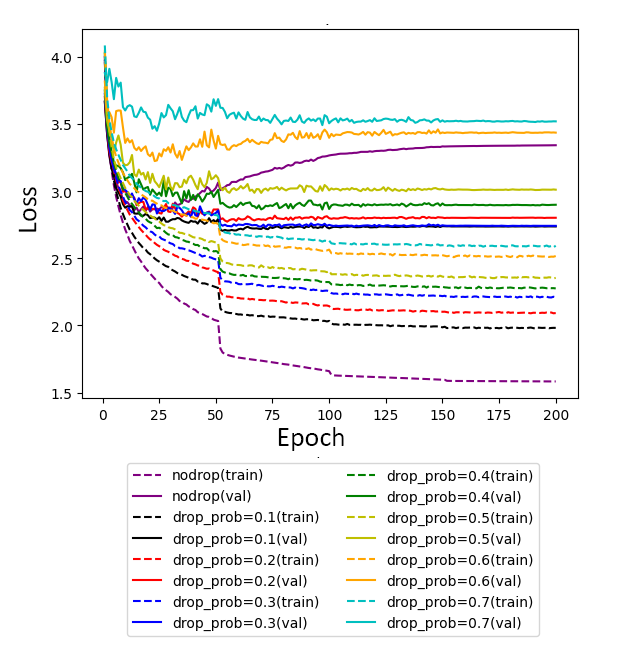} &
\includegraphics[width=\awidf]{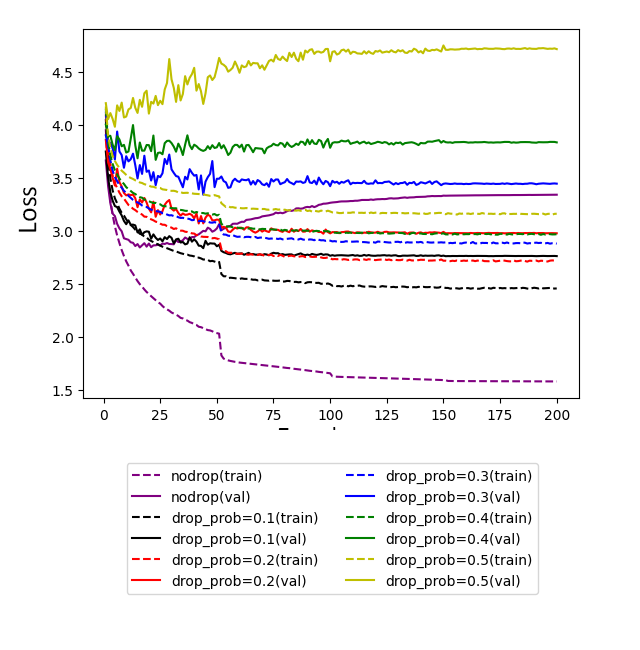} &
\includegraphics[width=\awidf]{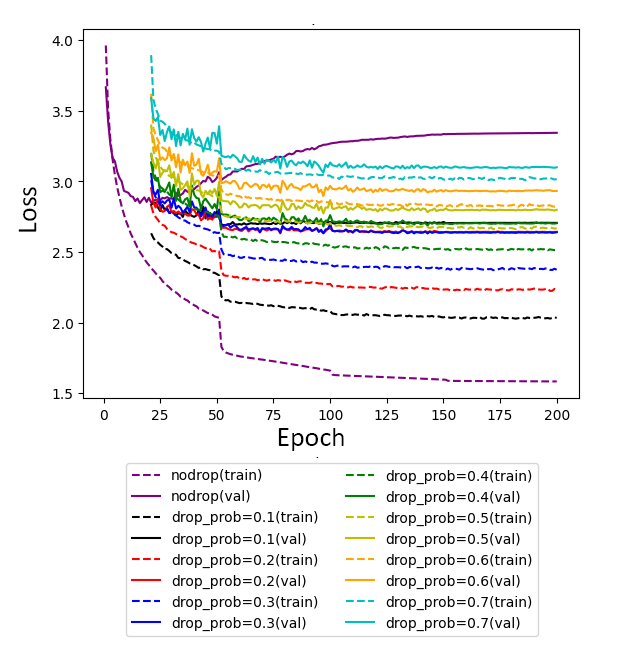} \\
(d) Dropout-CIFAR100-LeNet5 &
(e) DropBlock-CIFAR100-LeNet5 &
(f) DropCluster-CIFAR100-LeNet5\\


\includegraphics[width=\awidf]{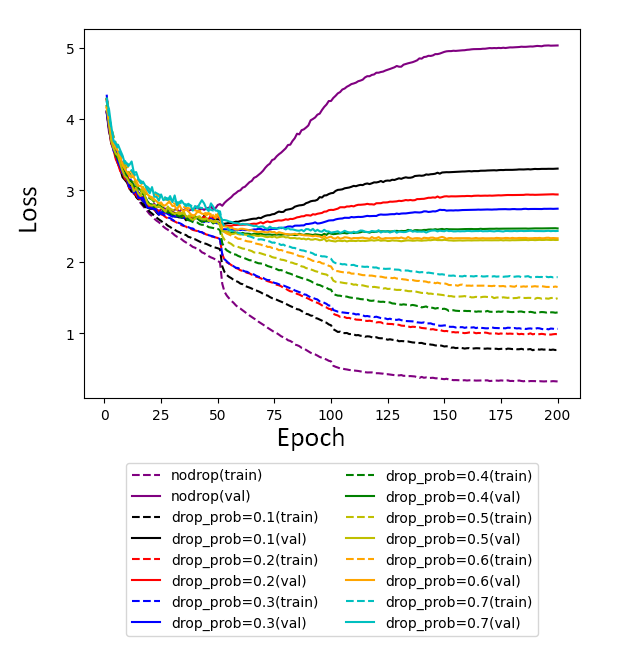} &
\includegraphics[width=\awidf]{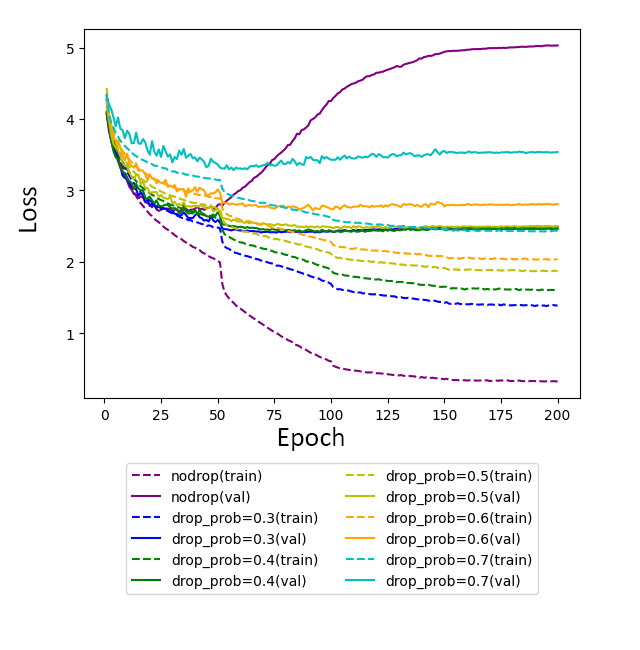} &
\includegraphics[width=\awidf]{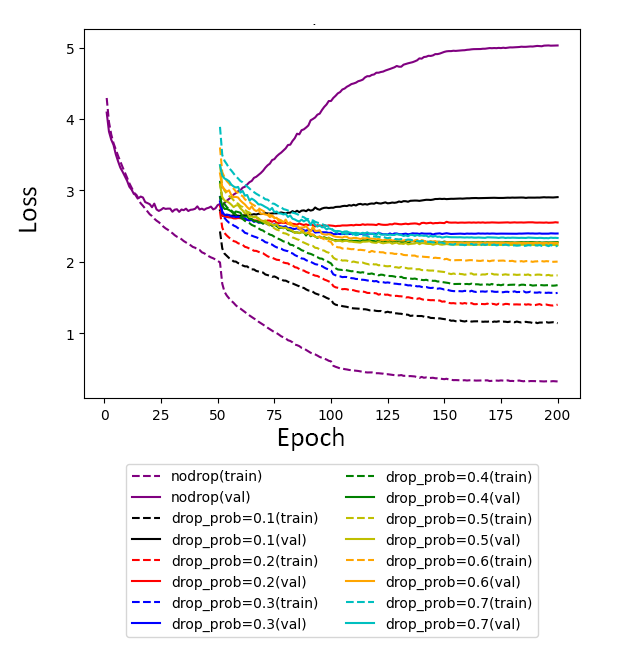} \\
(g) Dropout-CIFAR100-AlexNet &
(h) DropBlock-CIFAR100-AlexNet &
(i) DropCluster-CIFAR100-AlexNet

\end{tabular}
\caption{Ablation study on CIFAR-10/100 datasets.}
\label{fig:ablation_cifar}
\end{table*}

%
%

\begin{table*}[h]
\centering
\begin{tabular}{ccc}
 \includegraphics[width=\awidf]{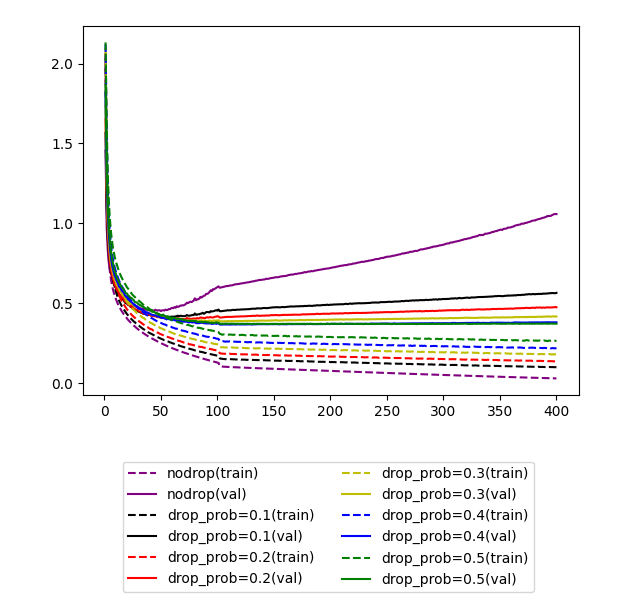} &
\includegraphics[width=\awidf]{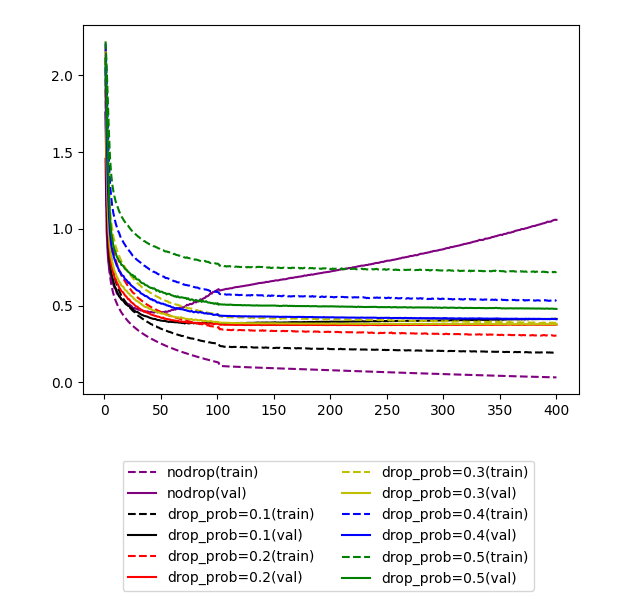} &
\includegraphics[width=\awidf]{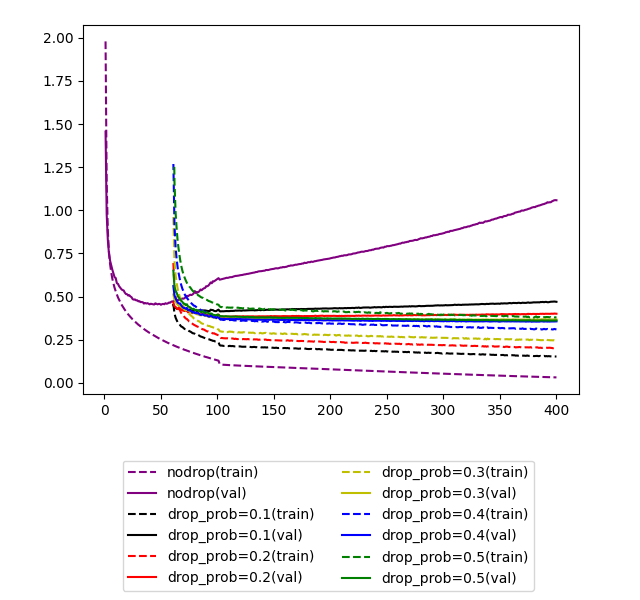} \\
(a) Dropout-SVHN-LeNet5 &
(b) DropBlock-SVHN-LeNet5 &
(c) DropCluster-SVHN-LeNet5 \\


\includegraphics[width=\awidf]{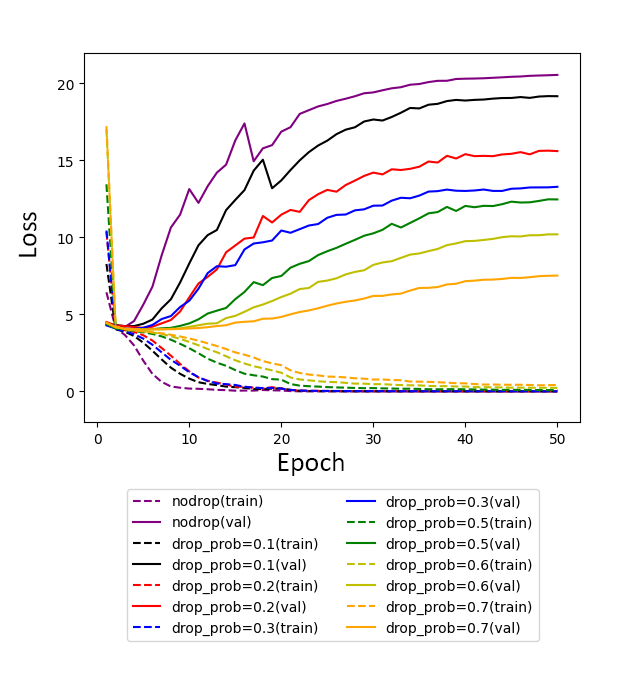} &
\includegraphics[width=\awidf]{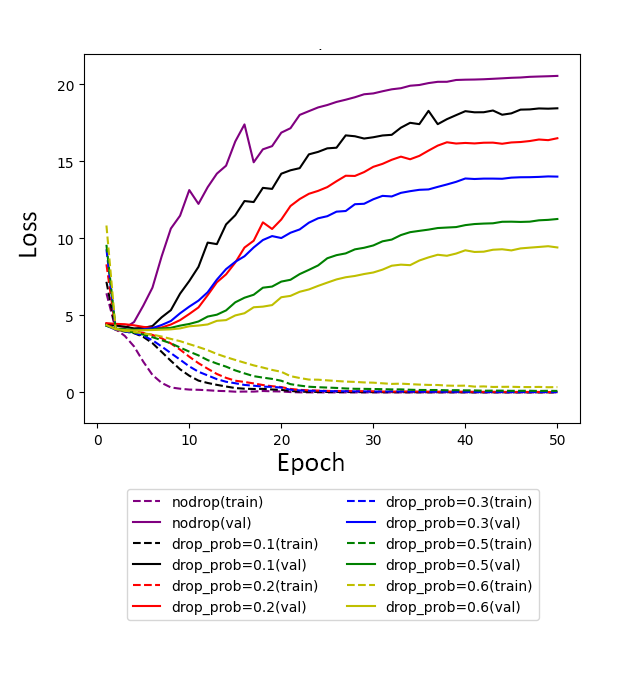} &
\includegraphics[width=\awidf]{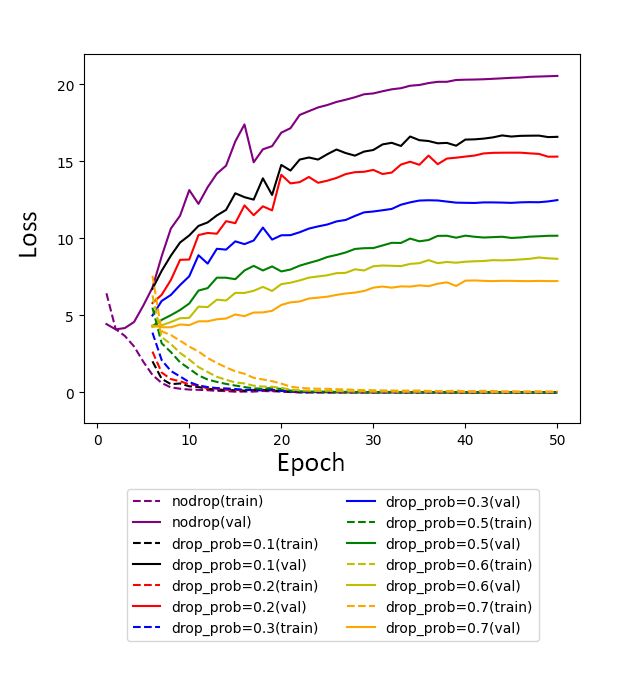} \\
(d) Dropout-APPA-CLF-LeNet5 &
(e) DropBlock-APPA-CLF-LeNet5 &
(f) DropCluster-APPA-CLF-LeNet5\\


\includegraphics[width=\awidf]{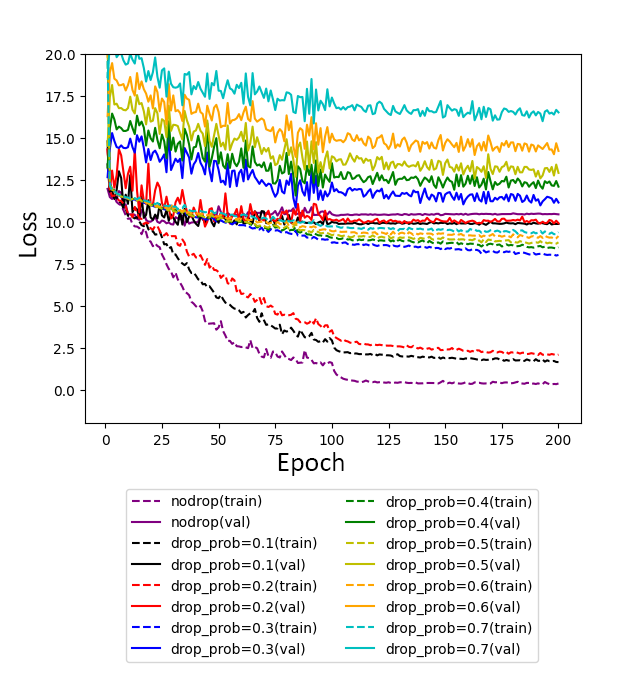} &
\includegraphics[width=\awidf]{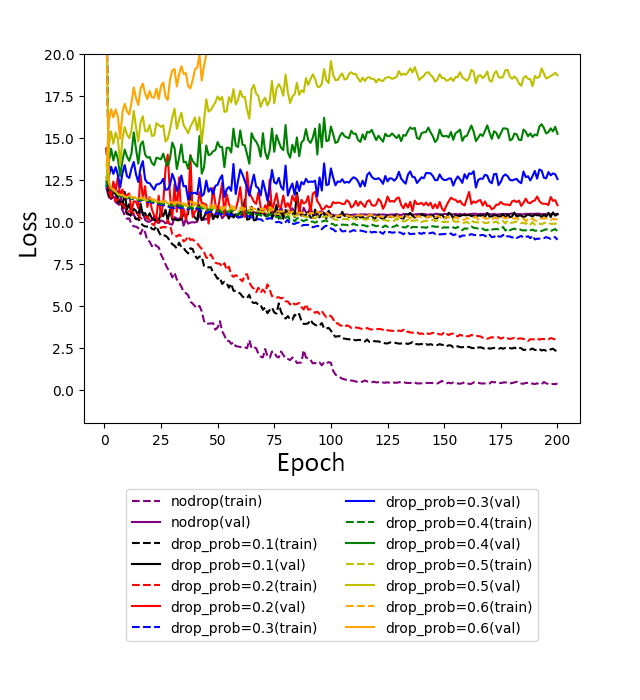} &
\includegraphics[width=\awidf]{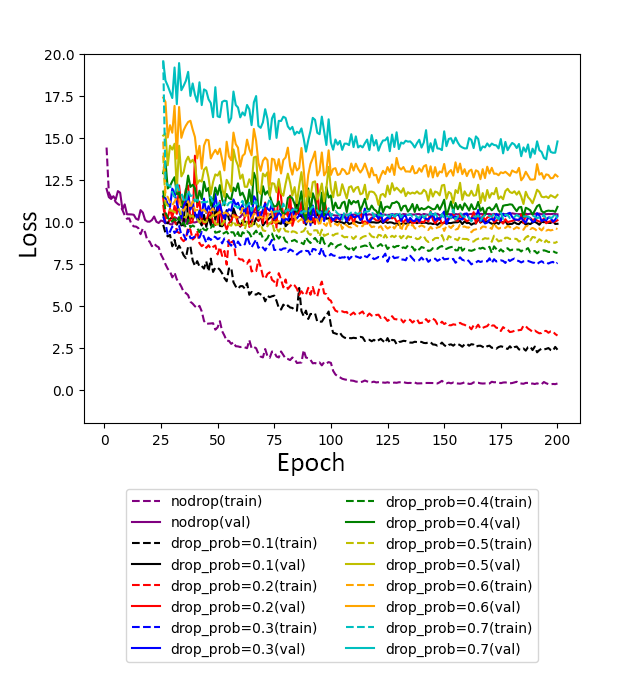} \\
(g) Dropout-APPA-REG-LeNet5 &
(h) DropBlock-APPA-REG-LeNet5 &
(i) DropCluster-APPA-REG-LeNet5

\end{tabular}
\caption{Ablation study on SVNH and APPA-REAL datasets.}
\label{fig:ablation_svnh_age}
\end{table*}

%% file: main.bbl
\begin{thebibliography}{10}
\providecommand{\url}[1]{#1}
\csname url@samestyle\endcsname
\providecommand{\newblock}{\relax}
\providecommand{\bibinfo}[2]{#2}
\providecommand{\BIBentrySTDinterwordspacing}{\spaceskip=0pt\relax}
\providecommand{\BIBentryALTinterwordstretchfactor}{4}
\providecommand{\BIBentryALTinterwordspacing}{\spaceskip=\fontdimen2\font plus
\BIBentryALTinterwordstretchfactor\fontdimen3\font minus \fontdimen4\font\relax}
\providecommand{\BIBforeignlanguage}[2]{{%
\expandafter\ifx\csname l@#1\endcsname\relax
\typeout{** WARNING: IEEEtran.bst: No hyphenation pattern has been}%
\typeout{** loaded for the language `#1'. Using the pattern for}%
\typeout{** the default language instead.}%
\else
\language=\csname l@#1\endcsname
\fi
#2}}
\providecommand{\BIBdecl}{\relax}
\BIBdecl

\bibitem{JMLR:v15:srivastava14a}
N.~Srivastava, G.~Hinton, A.~Krizhevsky, I.~Sutskever, and R.~Salakhutdinov, ``Dropout: A simple way to prevent neural networks from overfitting,'' \emph{Journal of Machine Learning Research}, vol.~15, pp. 1929--1958, 2014.

\bibitem{ghiasi2018dropblock}
G.~Ghiasi, T.-Y. Lin, and Q.~V. Le, ``Dropblock: A regularization method for convolutional networks,'' in \emph{Advances in Neural Information Processing Systems}, 2018, pp. 10\,727--10\,737.

\bibitem{mostafa2021visualizing}
S.~Mostafa, D.~Mondal, M.~Beck, C.~Bidinosti, C.~Henry, and I.~Stavness, ``Visualizing feature maps for model selection in convolutional neural networks,'' in \emph{IEEE/CVF International Conference on Computer Vision}, 2021, pp. 1362--1371.

\bibitem{tompson2015efficient}
J.~Tompson, R.~Goroshin, A.~Jain, Y.~LeCun, and C.~Bregler, ``Efficient object localization using convolutional networks,'' in \emph{IEEE/CVF Conference on Computer Vision and Pattern Recognition}, 2015, pp. 648--656.

\bibitem{fei2006one}
L.~Fei-Fei, R.~Fergus, and P.~Perona, ``One-shot learning of object categories,'' \emph{IEEE Transactions on Pattern Analysis and Machine Intelligence}, vol.~28, no.~4, pp. 594--611, 2006.

\bibitem{lake2011one}
B.~Lake, R.~Salakhutdinov, J.~Gross, and J.~Tenenbaum, ``One shot learning of simple visual concepts,'' in \emph{Proceedings of the Annual Meeting of the Cognitive Science Society}, vol.~33, 2011.

\bibitem{snell2017prototypical}
J.~Snell, K.~Swersky, and R.~Zemel, ``Prototypical networks for few-shot learning,'' \emph{Advances in Neural Information Processing Systems}, vol.~30, 2017.

\bibitem{liu2023dropout}
Z.~Liu, Z.~Xu, J.~Jin, Z.~Shen, and T.~Darrell, ``Dropout reduces underfitting,'' \emph{arXiv preprint arXiv:2303.01500}, 2023.

\bibitem{ravi2016optimization}
S.~Ravi and H.~Larochelle, ``Optimization as a model for few-shot learning,'' in \emph{International Conference on Learning Representations}, 2017.

\bibitem{kaiser2017learning}
L.~Kaiser, O.~Nachum, A.~Roy, and S.~Bengio, ``Learning to remember rare events,'' in \emph{International Conference on Learning Representations}, 2017.

\bibitem{fan2006statistical}
J.~Fan and R.~Li, ``Statistical challenges with high dimensionality: Feature selection in knowledge discovery,'' in \emph{Proceedings of the International Congress of Mathematicians}, 2006, pp. 595--622.

\bibitem{10002015global}
{1000 Genomes Project Consortium}, ``A global reference for human genetic variation,'' \emph{Nature}, vol. 526, no. 7571, p.~68, 2015.

\bibitem{huang2016deep}
G.~Huang, Y.~Sun, Z.~Liu, D.~Sedra, and K.~Q. Weinberger, ``Deep networks with stochastic depth,'' in \emph{European Conference on Computer Vision}, 2016, pp. 646--661.

\bibitem{DBLP:conf/aaai/HouW19}
S.~Hou and Z.~Wang, ``Weighted channel dropout for regularization of deep convolutional neural network,'' in \emph{Association for the Advancement of Artificial Intelligence}, 2019, pp. 8425--8432.

\bibitem{dai2019batch}
Z.~Dai, M.~Chen, X.~Gu, S.~Zhu, and P.~Tan, ``Batch dropblock network for person re-identification and beyond,'' in \emph{IEEE/CVF International Conference on Computer Vision}, 2019, pp. 3691--3701.

\bibitem{DBLP:conf/ijcai/ChenNLT20}
Z.~Chen, J.~Niu, X.~Liu, and S.~Tang, ``Selectscale: Mining more patterns from images via selective and soft dropout,'' in \emph{International Joint Conference on Artificial Intelligence}, 2020, pp. 523--529.

\bibitem{DBLP:conf/icml/ShiZDZMW20}
B.~Shi, D.~Zhang, Q.~Dai, Z.~Zhu, Y.~Mu, and J.~Wang, ``Informative dropout for robust representation learning: {A} shape-bias perspective,'' in \emph{International Conference on Machine Learning}, 2020, pp. 8828--8839.

\bibitem{DBLP:journals/ijcv/ZuninoBMZSM21}
A.~Zunino, S.~A. Bargal, P.~Morerio, J.~Zhang, S.~Sclaroff, and V.~Murino, ``Excitation dropout: Encouraging plasticity in deep neural networks,'' \emph{International Journal of Computer Vision}, vol. 129, no.~4, pp. 1139--1152, 2021.

\bibitem{DBLP:conf/aaai/PhamL21}
H.~Pham and Q.~V. Le, ``Autodropout: Learning dropout patterns to regularize deep networks,'' in \emph{Association for the Advancement of Artificial Intelligence}, 2021, pp. 9351--9359.

\bibitem{pmlr-v28-wan13}
L.~Wan, M.~Zeiler, S.~Zhang, Y.~Le~Cun, and R.~Fergus, ``Regularization of neural networks using dropconnect,'' in \emph{International Conference on Machine Learning}, ser. Proceedings of Machine Learning Research.\hskip 1em plus 0.5em minus 0.4em\relax PMLR, 2013, pp. 1058--1066.

\bibitem{DBLP:conf/cvpr/PalLVH20}
A.~Pal, C.~Lane, R.~Vidal, and B.~D. Haeffele, ``On the regularization properties of structured dropout,'' in \emph{IEEE/CVF Conference on Computer Vision and Pattern Recognition}, 2020, pp. 7668--7676.

\bibitem{zeng2020corrdrop}
Y.~Zeng, T.~Dai, and S.-T. Xia, ``Corrdrop: Correlation based dropout for convolutional neural networks,'' in \emph{ICASSP 2020-2020 IEEE International Conference on Acoustics, Speech and Signal Processing (ICASSP)}.\hskip 1em plus 0.5em minus 0.4em\relax IEEE, 2020, pp. 3742--3746.

\bibitem{liu2023patchdropout}
Y.~Liu, C.~Matsoukas, F.~Strand, H.~Azizpour, and K.~Smith, ``Patchdropout: Economizing vision transformers using patch dropout,'' in \emph{IEEE Winter Conference on Applications of Computer Vision}, 2023, pp. 3953--3962.

\bibitem{lee2022self}
H.~Lee, Y.~Park, H.~Seo, and M.~Kang, ``Self-knowledge distillation via dropout,'' \emph{arXiv preprint arXiv:2208.05642}, 2022.

\bibitem{lin2023explore}
S.~Lin, X.~Zeng, and R.~Zhao, ``Explore the power of dropout on few-shot learning,'' \emph{arXiv preprint arXiv:2301.11015}, 2023.

\bibitem{zhang2019confidence}
Z.~Zhang, A.~V. Dalca, and M.~R. Sabuncu, ``Confidence calibration for convolutional neural networks using structured dropout,'' \emph{arXiv preprint arXiv:1906.09551}, 2019.

\bibitem{veit2016residual}
A.~Veit, M.~J. Wilber, and S.~Belongie, ``Residual networks behave like ensembles of relatively shallow networks,'' 2016.

\bibitem{kim2023use}
B.~J. Kim, H.~Choi, H.~Jang, D.~Lee, and S.~W. Kim, ``How to use dropout correctly on residual networks with batch normalization,'' \emph{arXiv preprint arXiv:2302.06112}, 2023.

\bibitem{ba2013adaptive}
J.~Ba and B.~Frey, ``Adaptive dropout for training deep neural networks,'' \emph{Advances in Neural Information Processing Systems}, vol.~26, 2013.

\bibitem{dodballapur2020automatic}
V.~Dodballapur, R.~Calisa, Y.~Song, and W.~Cai, ``Automatic dropout for deep neural networks,'' in \emph{International Conference on Neural Information Processing}, 2020, pp. 185--196.

\bibitem{fan2021contextual}
X.~Fan, S.~Zhang, K.~Tanwisuth, X.~Qian, and M.~Zhou, ``Contextual dropout: An efficient sample-dependent dropout module,'' in \emph{International Conference on Learning Representations}, 2021.

\bibitem{li2022dropkey}
B.~Li, Y.~Hu, X.~Nie, C.~Han, X.~Jiang, T.~Guo, and L.~Liu, ``Dropkey,'' \emph{arXiv preprint arXiv:2208.02646}, 2022.

\bibitem{banerjee2004validating}
A.~Banerjee and R.~N. Dave, ``Validating clusters using the {Hopkins} statistic,'' in \emph{IEEE International Conference on Fuzzy Systems}, 2004, pp. 149--153.

\bibitem{lecun1989backpropagation}
Y.~LeCun, B.~Boser, J.~S. Denker, D.~Henderson, R.~E. Howard, W.~Hubbard, and L.~D. Jackel, ``Backpropagation applied to handwritten zip code recognition,'' \emph{Neural Computation}, vol.~1, no.~4, pp. 541--551, 1989.

\bibitem{krizhevsky2009learning}
A.~Krizhevsky, ``Learning multiple layers of features from tiny images,'' Master's Thesis, University of Toronto, 2009.

\bibitem{lloyd1982least}
S.~Lloyd, ``Least squares quantization in pcm,'' \emph{IEEE Transactions on Information Theory}, vol.~28, no.~2, pp. 129--137, 1982.

\bibitem{rousseeuw1987silhouettes}
P.~J. Rousseeuw, ``Silhouettes: a graphical aid to the interpretation and validation of cluster analysis,'' \emph{Journal of Computational and Applied Mathematics}, vol.~20, pp. 53--65, 1987.

\bibitem{cavazza2018dropout}
J.~Cavazza, P.~Morerio, B.~Haeffele, C.~Lane, V.~Murino, and R.~Vidal, ``Dropout as a low-rank regularizer for matrix factorization,'' in \emph{International Conference on Artificial Intelligence and Statistics}.\hskip 1em plus 0.5em minus 0.4em\relax PMLR, 2018, pp. 435--444.

\bibitem{mianjy2018implicit}
P.~Mianjy, R.~Arora, and R.~Vidal, ``On the implicit bias of dropout,'' in \emph{International Conference on Machine Learning}, 2018, pp. 3540--3548.

\bibitem{krizhevsky2012imagenet}
A.~Krizhevsky, I.~Sutskever, and G.~E. Hinton, ``Imagenet classification with deep convolutional neural networks,'' in \emph{Advances in Neural Information Processing Systems}, 2012, pp. 1097--1105.

\bibitem{netzer2011reading}
Y.~Netzer, T.~Wang, A.~Coates, A.~Bissacco, B.~Wu, and A.~Y. Ng, ``Reading digits in natural images with unsupervised feature learning,'' in \emph{Advances in Neural Information Processing Systems}, 2011.

\bibitem{clapes2018apparent}
A.~Clap{\'e}s, O.~Bilici, D.~Temirova, E.~Avots, G.~Anbarjafari, and S.~Escalera, ``From apparent to real age: gender, age, ethnic, makeup, and expression bias analysis in real age estimation,'' in \emph{IEEE Conference on Computer Vision and Pattern Recognition Workshops}, 2018, pp. 2373--2382.

\bibitem{dropblock_github}
M.~V. Ramos, ``Dropblock,'' \url{https://github.com/miguelvr/dropblock}, 2019, [Online; accessed February 2020].

\bibitem{murphy2012machine}
K.~P. Murphy, \emph{Machine learning: a probabilistic perspective}.\hskip 1em plus 0.5em minus 0.4em\relax MIT press, 2012.

\bibitem{kingma2014adam}
D.~P. Kingma and J.~Ba, ``Adam: A method for stochastic optimization,'' in \emph{International Conference on Learning Representations}, Y.~Bengio and Y.~LeCun, Eds., 2015.

\bibitem{oliphant2007python}
T.~E. Oliphant, ``Python for scientific computing,'' \emph{Computing in Science \& Engineering}, vol.~9, no.~3, 2007.

\bibitem{paszke2019pytorch}
A.~Paszke, S.~Gross, F.~Massa \emph{et~al.}, ``Pytorch: An imperative style, high-performance deep learning library,'' in \emph{Advances in Neural Information Processing Systems}, 2019, pp. 8024--8035.

\bibitem{pedregosa2011scikit}
F.~Pedregosa, G.~Varoquaux, A.~Gramfort \emph{et~al.}, ``Scikit-learn: Machine learning in python,'' \emph{Journal of Machine Learning Research}, vol.~12, pp. 2825--2830, 2011.

\bibitem{walt2011numpy}
S.~v.~d. Walt, S.~C. Colbert, and G.~Varoquaux, ``The numpy array: a structure for efficient numerical computation,'' \emph{Computing in Science \& Engineering}, vol.~13, no.~2, pp. 22--30, 2011.

\end{thebibliography}
